\ifdefined\pdfsuppressptexinfo
\fi
\documentclass{article}

\usepackage{xcolor}         
\usepackage{hyperref}

\usepackage[accepted]{mlsys2025}
\renewcommand{\toptitlebar}{\hrule height0pt \vskip .25in}
\renewcommand{\bottomtitlebar}{\vskip .22in \hrule height0pt \vskip .3in}
\mlsystitlerunning{\rorekv: Accelerating LLM Inference Serving through Uncertainty and Redundancy Aware KV Cache Compression}

\usepackage{mathtools}
\usepackage{standalone}
\usepackage{tikz}
\usepackage{graphicx}
\usetikzlibrary{decorations.pathreplacing,positioning}
\usepackage{amsmath}
\usepackage[utf8]{inputenc} 
\usepackage[T1]{fontenc}    
\usepackage{hyperref}       
\usepackage{url}            
\usepackage{booktabs}       
\usepackage{amsfonts}       
\usepackage{nicefrac}       
\usepackage{microtype}      

\usepackage{amsmath} 
\usepackage{multirow} 
\usepackage{caption}
\usepackage{microtype}
\usepackage{graphicx}
\usepackage{subfigure}

\usepackage{booktabs} 
\usepackage{hyperref}
\usepackage{multirow}
\usepackage{algorithm}
\usepackage{algorithmic}
\usepackage{xspace}

\newcommand{\rocketkv}{{RocketKV}\xspace}
\newcommand{\rorekv}{{RoCoKV}\xspace}

\newcommand{\snapkv}{{SnapKV}\xspace}
\newcommand{\fullkv}{{Full-KV}\xspace}

\newcommand{\rkv}{{R-KV}\xspace}
\newcommand{\raas}{{RaaS}\xspace}

\newcommand{\topk}{{top-\it{k}}\xspace}
\newcommand{\topp}{{top-\it{p}}\xspace}
\newcommand{\passone}{{pass@1}\xspace}
\newcommand{\vllm}{{vLLM}\xspace}
\newcommand{\deepllama}
{{R1-Llama-8B}\xspace}
\newcommand{\deepqwen}
{{R1-Qwen-7B}\xspace}
\newcommand{\gsm}{{GSM8K}\xspace}
\newcommand{\mathf}{{MATH500}\xspace}
\newcommand{\aime}{{AIME25}\xspace}

\newcommand{\needle}
{{Needle-in-a-Haystack}\xspace}

\begin{document}
\twocolumn[
\mlsystitle{\rorekv: Accelerating LLM Inference Serving through Uncertainty and Redundancy Aware KV Cache Compression}

\mlsyssetsymbol{equal}{*}

\begin{mlsysauthorlist}
\mlsysauthor{Payman Behnam}{equal,to}
\mlsysauthor{Elaheh Sobhani}{equal,goo}
\mlsysauthor{Saad Amin}{to}
\mlsysauthor{Yahya Hassan}{ama}
\mlsysauthor{Daniyal Khan}{to}
\mlsysauthor{Biswadeep Chakraborty}{ed}
\mlsysauthor{Tanish Patwa}{foo}
\mlsysauthor{Souvik Kundu}{int}
\mlsysauthor{Alexey Tumanov}{to}
\end{mlsysauthorlist}

\mlsysaffiliation{to}{Georgia Institute of Technology, USA}
\mlsysaffiliation{goo}{Independent Researcher, USA}
\mlsysaffiliation{ama}{Annapurna Labs, Amazon, USA; work done while at Georgia Institute of Technology.}
\mlsysaffiliation{ed}{Mercuria Energy Americas, USA; work done while at Georgia Institute of Technology.}
\mlsysaffiliation{foo}{Google, USA; work done while at Georgia Institute
of Technology.}
\mlsysaffiliation{int}{Intel, USA}

\mlsyscorrespondingauthor{Payman Behnam}{payman.behnam@gatech.edu}
\mlsyscorrespondingauthor{Alexey Tumanov}{atumanov@gatech.edu}

\mlsyskeywords{Machine Learning, MLSys}

\vskip 0.3in

\label{sec:0-abstract}
\begin{abstract}

KV cache compression is critical for large reasoning models, where most of the cost comes from long generated reasoning traces rather than the input prompt. These traces often contain redundant and low-utility tokens due to looping, self-verification, and corrections. However, such tokens may still receive high attention scores, making it difficult to remove them without harming the reasoning process. We propose \rorekv, a KV cache compression method designed for large reasoning models. \rorekv introduces two key components. First, uncertainty-guided adaptive pruning adjusts the compression strength based on the model’s confidence, allocating more memory during uncertain phases and applying stronger pruning when the model is confident. Second, redundancy-aware token selection reduces the impact of tokens that are highly similar to recent reasoning content, encouraging the model to retain more informative and diverse tokens.
Beyond the algorithm, we co-design \rorekv with the serving stack: we develop fused three-phase CUDA decode kernels and integrate them into \vllm as a drop-in plugin that preserves its paged KV layout and CUDA graph capture, turning algorithmic savings into real end-to-end speedups.
Extensive experiments across three reasoning benchmarks and two reasoning models demonstrate that \rorekv consistently improves reasoning accuracy under constrained KV cache budgets, achieving higher accuracy compared to state-of-the-art KV cache compression methods. 
For DeepSeek-R1-Distill-Llama-8B in \vllm, \rorekv achieves a 1.27$\times$ speedup and 38\% higher throughput on \aime, the most challenging evaluated reasoning benchmark, and a 1.43$\times$ speedup and 45\% higher throughput on the long synthetic benchmark, compared with \fullkv. 
\end{abstract}]
\printAffiliationsAndNotice{\mlsysEqualContribution}
\section{Introduction}
\label{sec:1-intro}

KV cache compression becomes more difficult for large reasoning models (LRMs) because the cost (e.g., latency, memory, throughput) often comes from the long generated reasoning traces, not from the input prompt.  These generated tokens often arise from overthinking, looping, self-verification, corrections, etc. Hence, the KV cache may store a lot of non-useful information. These tokens may still receive high attention scores because they are temporally or semantically close to important tokens. As a result, attention-based selection methods may allocate budget to near-duplicate or low-utility content. Moreover, removing the wrong tokens can disrupt the chain of thought (CoT), introduce repetitive loops, or discard intermediate reasoning steps that may become useful later~\cite{rkv2025,headkvr22025,  seerattentionr2025, TriAttention2026, rpc2025, skipkv2026}.
This makes KV cache compression for reasoning models fundamentally different from non-reasoning settings, where token importance is typically more stable and easier to estimate from the input context~\cite{  rocketkv2025,magicpig2024, snapkv2024,shadowkv25,razor2024,quest2024}. Several recent works have explored KV cache compression for reasoning models. For instance, \rkv~\cite{rkv2025} targets the redundancy that appears in the long reasoning decode stage. Its key idea is to score tokens using both attention-based importance and key-vector-based redundancy, then keep tokens that are important but not repetitive during decoding. \raas ~\cite{raas2025} observes that reasoning traces contain intermediate milestone tokens. These tokens become important for a while and then fade away. \raas keeps such tokens using an LRU-style policy and preserves all prefill tokens to avoid losing tokens that may become important again later~\cite{raas2025}. SkipKV~\cite{skipkv2026} combines sentence-level KV cache eviction of semantically redundant reasoning with adaptive activation steering to discourage redundant generation. It further groups requests by prompt length to reduce padding and preserve the effective KV cache budget during batched inference. Overall, KV cache compression for reasoning models must not only reduce memory and latency but also preserve the structure and continuity of the model's reasoning trajectory.

From one perspective, KV cache compression for the decode stage can be divided into two groups: permanent KV cache eviction and query-aware KV selection. Permanent eviction methods, such as H2O~\cite{h2o2024} and \snapkv~\cite{snapkv2024}, reduce both KV cache storage and memory bandwidth. However, because the removed tokens cannot be recovered later, these methods usually suffer from noticeable accuracy degradation. Query-aware methods, such as Quest~\cite{quest2024}, SparQ~\cite{sparq2024}, and Loki~\cite{loki2024} address the accuracy drop issue by taking into account selecting important KV tokens based on the current query and approximating the attention computation more carefully. While  this improves accuracy, these methods usually do not reduce the total KV cache storage. In contrast, recent methods such as \rocketkv~\cite{rocketkv2025} combine permanent KV cache eviction with dynamic KV token selection. This allows them to maintain high accuracy even under small token budgets, while reducing both memory bandwidth and KV cache storage.

\rocketkv~\cite{rocketkv2025} employs a hybrid sparse attention (HSA) mechanism during decoding, where fine-grained \topk attention is approximated through reductions along both head and sequence dimensions. Although this design enables efficient sparse attention under fixed budgets, it does not explicitly account for the characteristics of long reasoning traces. In particular, it treats all tokens uniformly based on approximate attention scores. As a result, it may retain redundant intermediate CoT tokens and allocate budget inefficiently across decoding steps, especially in long and complex reasoning trajectories.

To address these limitations, we propose \rorekv, which extends existing KV compression methods with mechanisms tailored for reasoning workloads. First, we introduce uncertainty-guided adaptive pruning. The reasoning trajectory evolves with varying levels of confidence across decoding steps. During uncertain phases, the model requires more context to explore alternatives or verify intermediate results, while during confident phases, stronger pruning can be applied safely. We capture this behavior by dynamically adjusting the pruning strength based on a token-level uncertainty signal. Second, we introduce a redundancy-aware token scoring mechanism. The key idea is to reduce the contribution of tokens that are highly similar to previously generated reasoning content. This is achieved by penalizing their attention scores based on semantic similarity, encouraging the model to retain more diverse and informative tokens.

Moreover, a key limitation of many KV cache compression methods, such as \rocketkv, is that their reported gains are often measured at the algorithmic level, while their benefit in a production inference engine is less clear. Recent analyses show that reducing the KV footprint does not necessarily translate into end-to-end improvements in latency or throughput. In practice, compression introduces additional overheads such as scoring, eviction, and memory management, which can offset the benefits of reduced KV size. For example, a recent work~\cite{rethinkingserving2025} shows that H2O~\cite{h2o2024} reduces prefill and decode throughput to only $0.58\times$ and $0.69\times$ of the FP16 baseline under tensor parallelism. These observations suggest that KV compression methods on the algorithmic side should be designed along with the system-side and low-level hardware side. These bodies of work further observe that the speedups from KV compression diminish when combined with optimized serving primitives such as PagedAttention~\cite{pagedeviction2026}, making it difficult to integrate efficiently into the \vllm~\cite{vllm2023}. To bridge this gap, we co-design \rorekv with the serving stack. We build fused three-phase CUDA decode kernels that compute approximate scoring, redundancy, and exact sparse attention in a single pass over the KV cache, and integrate them into \vllm as a drop-in plugin that preserves the engine's paged KV layout. The entire pipeline, including per-request adaptive budgets, runs inside a single captured CUDA graph, translating algorithmic savings into practical end-to-end serving speedups.

Our contributions are summarized as follows:

\begin{itemize}

\item  We observe that longer reasoning traces are more likely to end with incorrect answers, so generating more tokens does not always help the model achieve better reasoning. We also observe that the effect of attention approximation depends strongly on the model's uncertainty. Similar errors in selecting cached tokens are more harmful when the model is uncertain than when it is confident. Hence, we should keep more context during uncertain steps and prune more during confident steps. They also motivate giving less attention to repeated content, so that the available attention budget is used for useful information.

\item Based on our observations, we propose \rorekv, which combines uncertainty-guided adaptive pruning with redundancy-aware token scoring. \rorekv selects more tokens during uncertain steps and prunes more strongly during confident steps, using the previous step's output distribution without an extra forward pass. Its redundancy score uses a compact summary of recent context to favor informative tokens while avoiding expensive pairwise comparisons across the cache.

\item We develop a drop-in \vllm plugin with a three-phase CUDA pipeline for approximate scoring, top-$k$ selection, and exact sparse attention. Fusing redundancy computation with scoring avoids additional full-cache passes, while GPU-resident adaptive budgets preserve CUDA graph capture without CPU-to-GPU synchronization. The integration preserves \vllm's page tables, scheduler, and prefix caching without changing its source code. Our design provides a practical reference for integrating other KV selection and sparse-attention methods into \vllm.

\item Across three reasoning benchmarks and two models, \rorekv consistently outperforms all evaluated state-of-the-art compression methods under tight budgets. \rorekv also achieves notable efficiency improvements. When serving DeepSeek-R1-Distill-Llama-8B in \vllm{}, \rorekv{} achieves a 1.27$\times$ speedup and 38\% higher throughput on \aime{}  as the most complex evaluated dataset, and a 1.43$\times$ speedup and 45\% higher throughput on the long synthetic dataset, compared with \fullkv{}.
\end{itemize}

\section{Proposed Method}
\label{sec:3-proposed}

\subsection{Observations}
\label{sec:observations}

\begin{figure*}[!h]
    \centering
  \subfigure[\scriptsize Attention mass preserved by \rocketkv relative to exact top-$k$ across uncertainty deciles.\label{fig:attention_vs_uncertainty}]{  
    \includegraphics[width=0.24\textwidth]{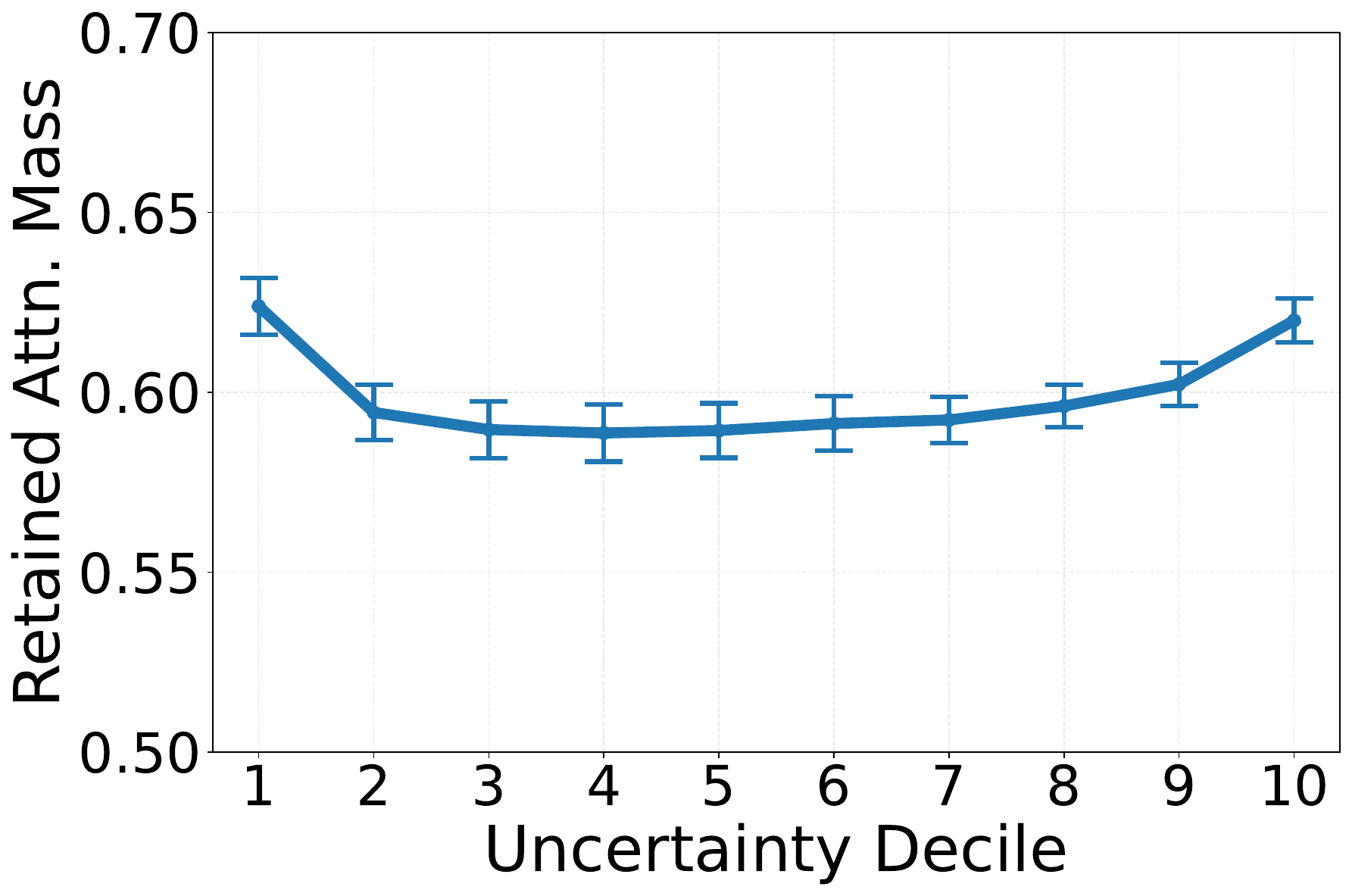}}
    \hfill
    \subfigure[\scriptsize Top-$k$ overlap: Fraction of keys shared by \rocketkv and exact top-$k$ selection across uncertainty deciles.\label{fig:overlap_vs_uncertainty}]{  
    \includegraphics[width=0.24\textwidth]{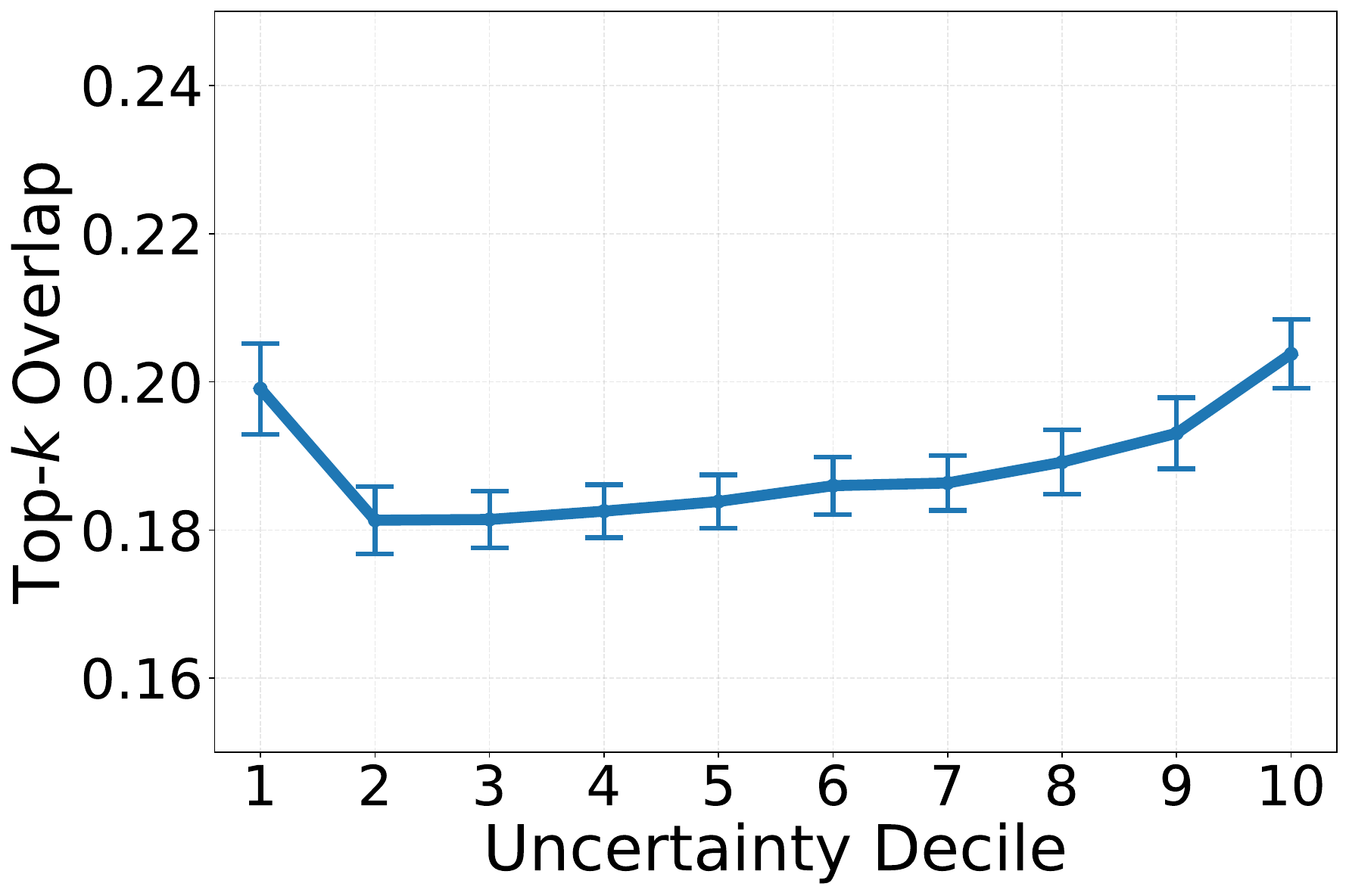}}
    \hfill
    \subfigure[\scriptsize Total variation: distance between next-token distributions produced by \rocketkv and exact top-$k$ selection.
\label{fig:total_variation}]{%
    \includegraphics[width=0.24\textwidth]{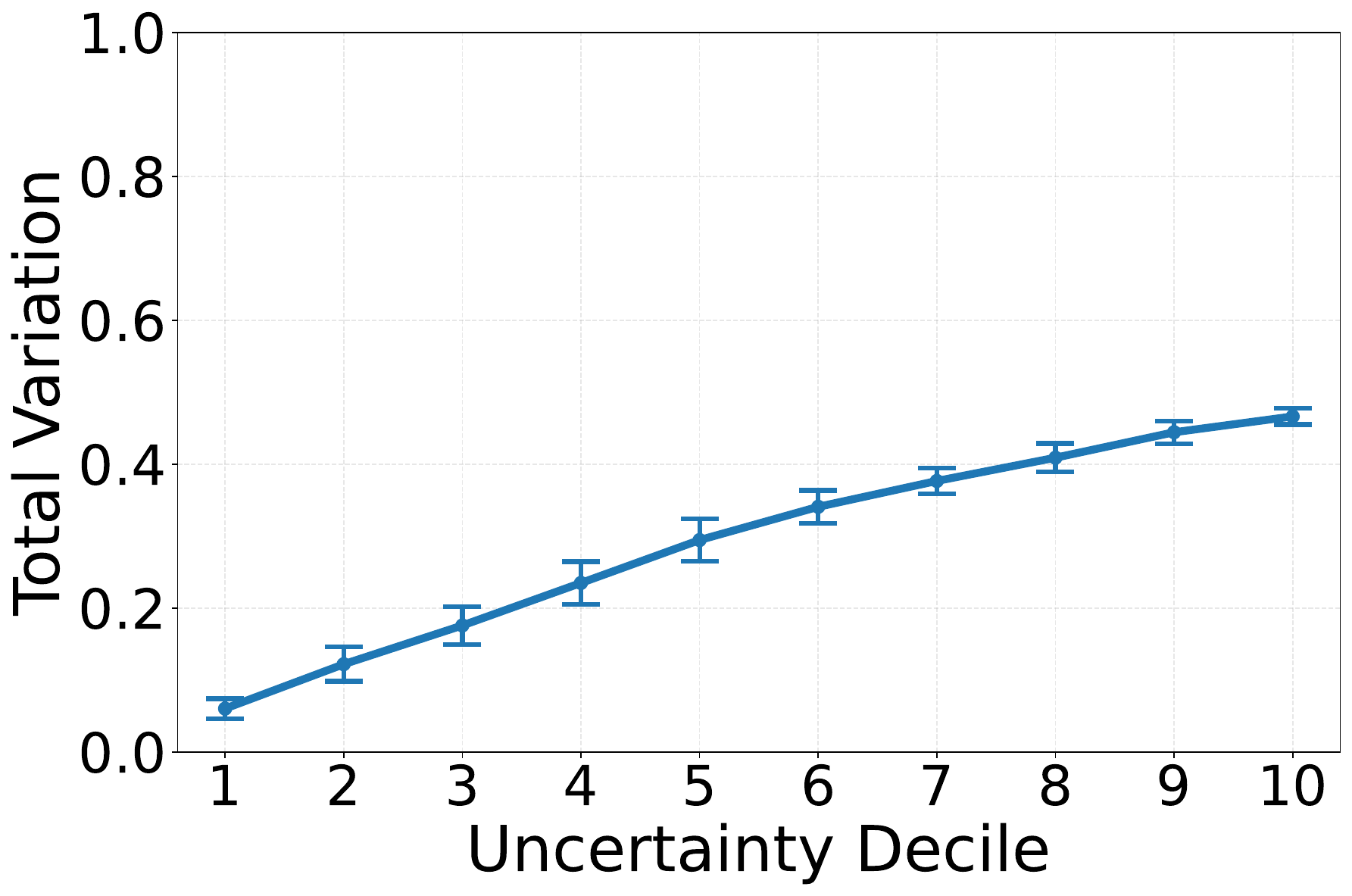}}
    \hfill
    \subfigure[\scriptsize Flip rate: fraction of steps where \rocketkv and exact top-$k$ select different next tokens from the same KV cache. \label{fig:flip_rate}]{\includegraphics[width=0.24\textwidth]{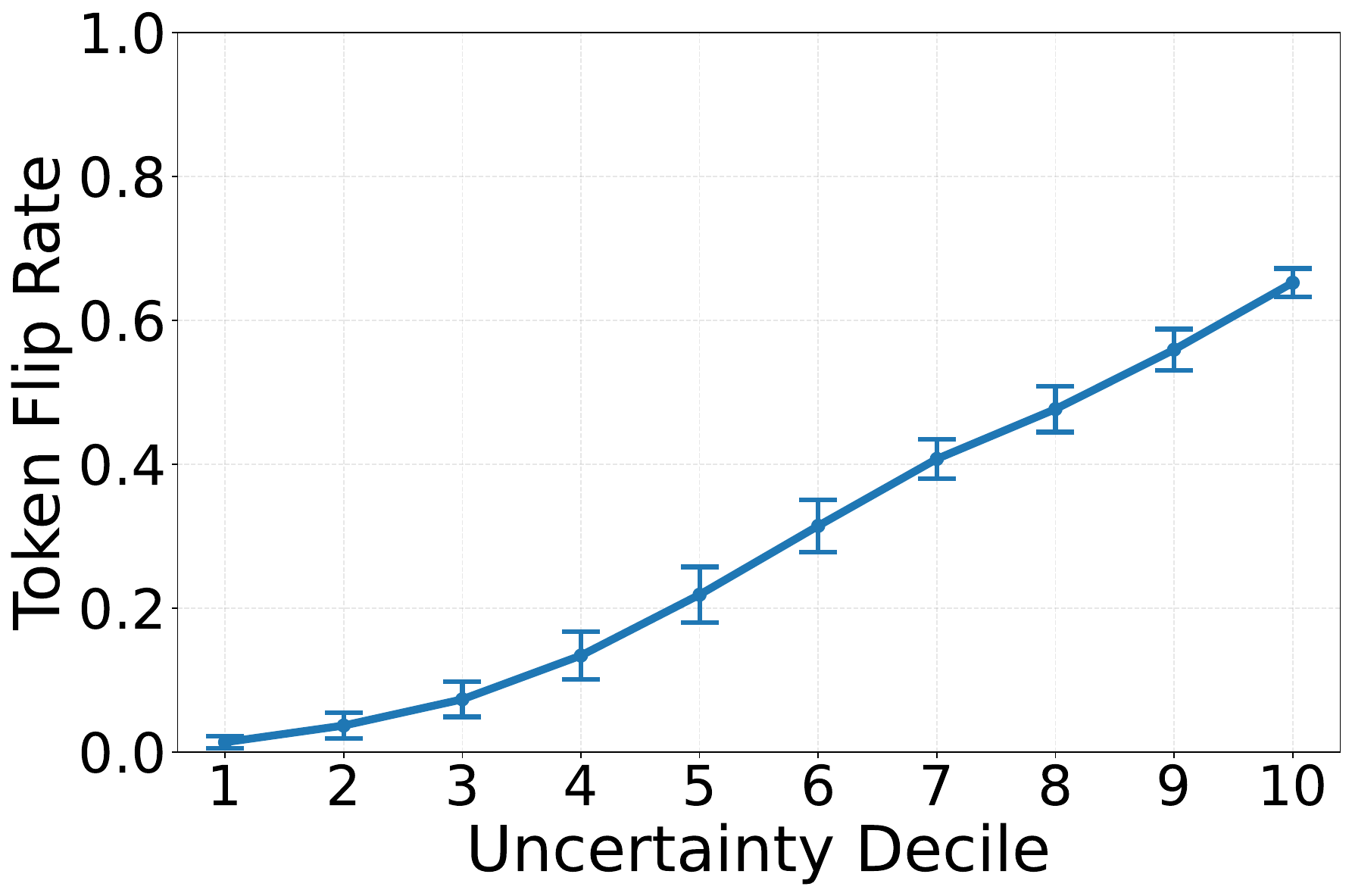}}
    
    \vspace{-2ex}
    \caption{Observation 1\,-\,Attention approximation errors are more harmful at uncertain decoding steps.}
    \label{fig:observation2}
\end{figure*}

\begin{figure*}[!ht]
    \centering
  \subfigure[\scriptsize \gsm, all answers\label{fig:gsm_all_answers}]{  
    \includegraphics[width=0.24\textwidth]{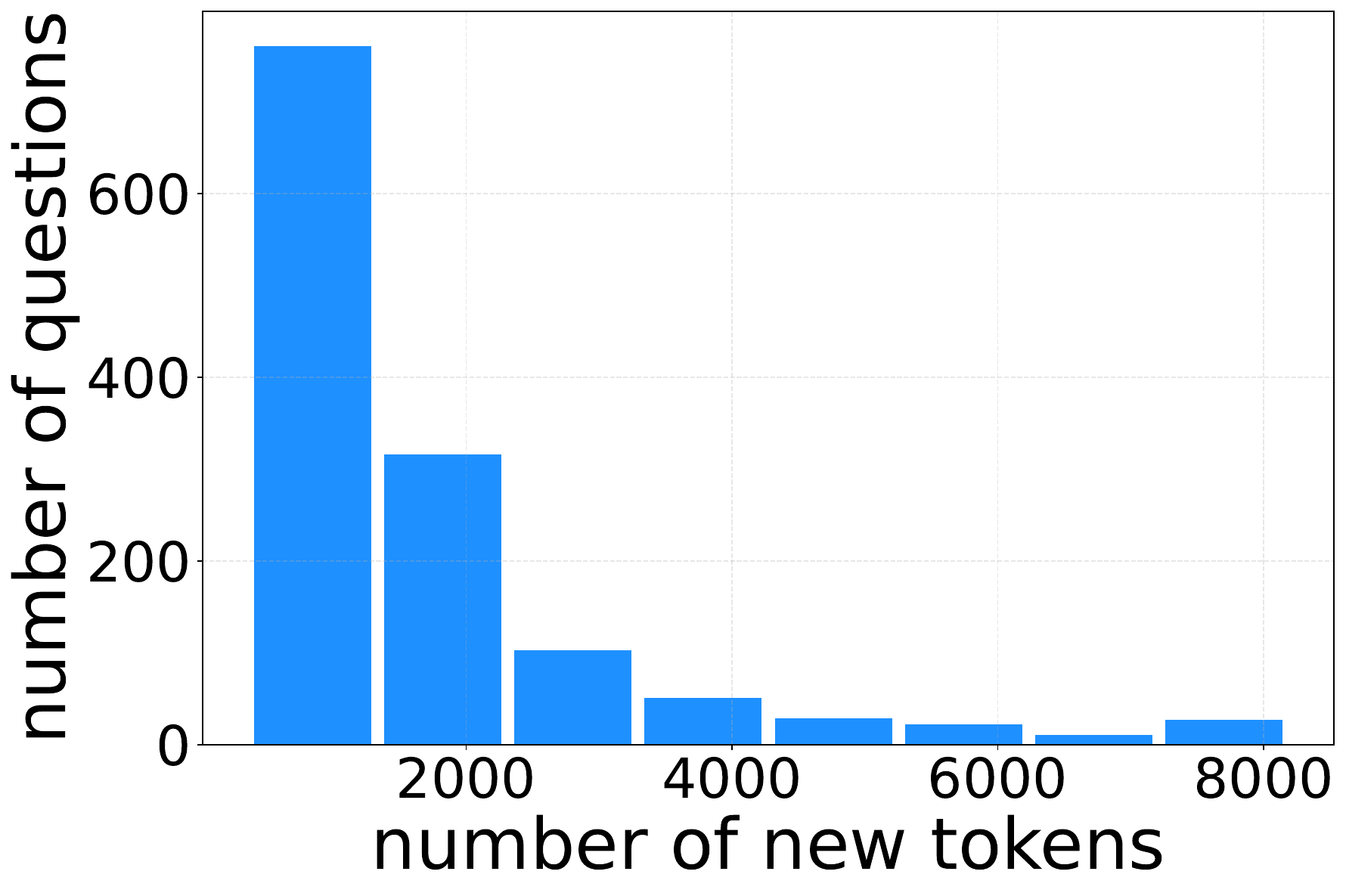}}
    \hfill
    \subfigure[\scriptsize \gsm, correct vs. wrong\label{fig:gsm_correct_wrong}]{  
    \includegraphics[width=0.24\textwidth]{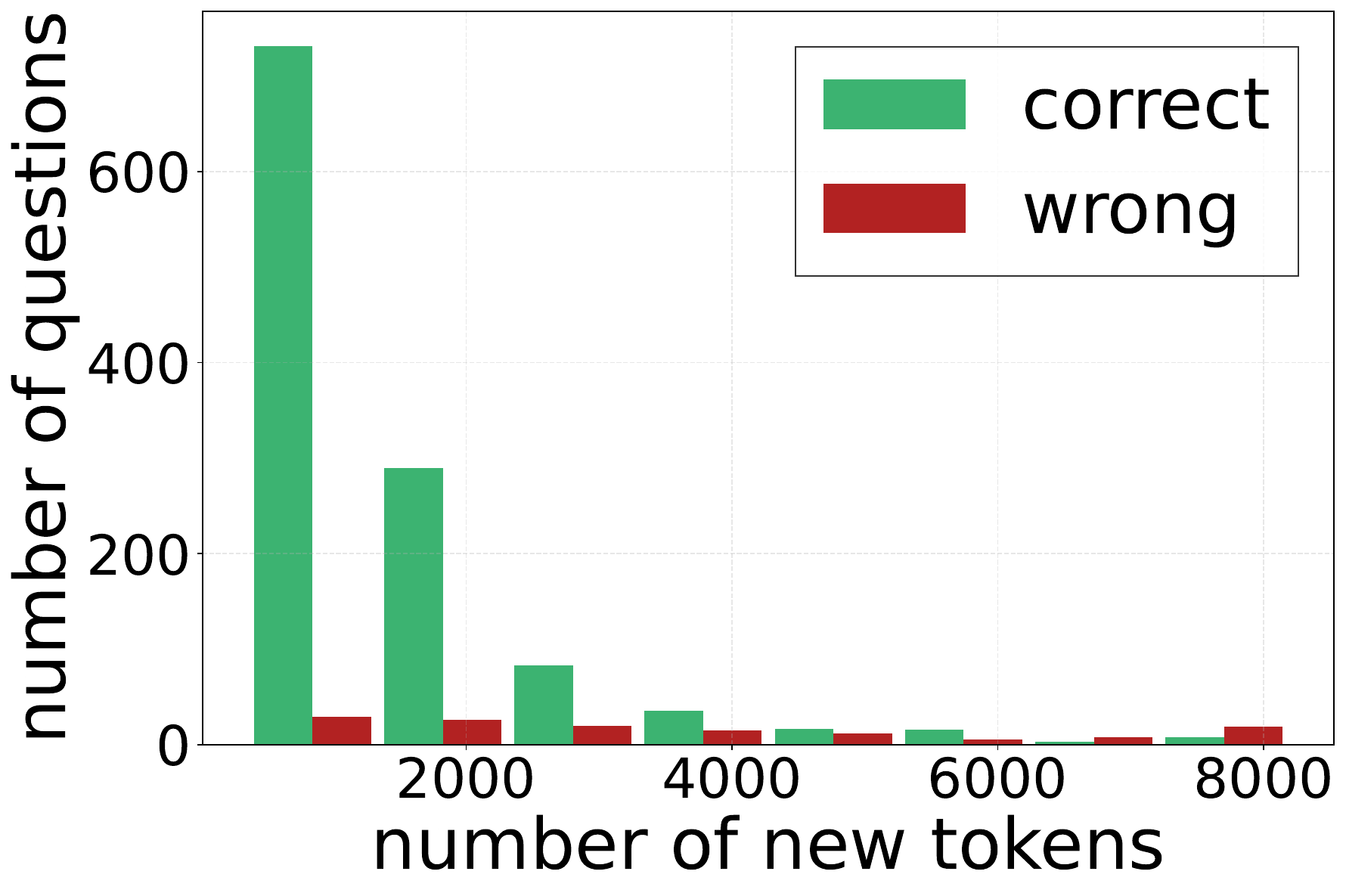}}
    \hfill
    \subfigure[\scriptsize \aime, all answers \label{fig:aime_all_answers}]{%
    \includegraphics[width=0.24\textwidth]{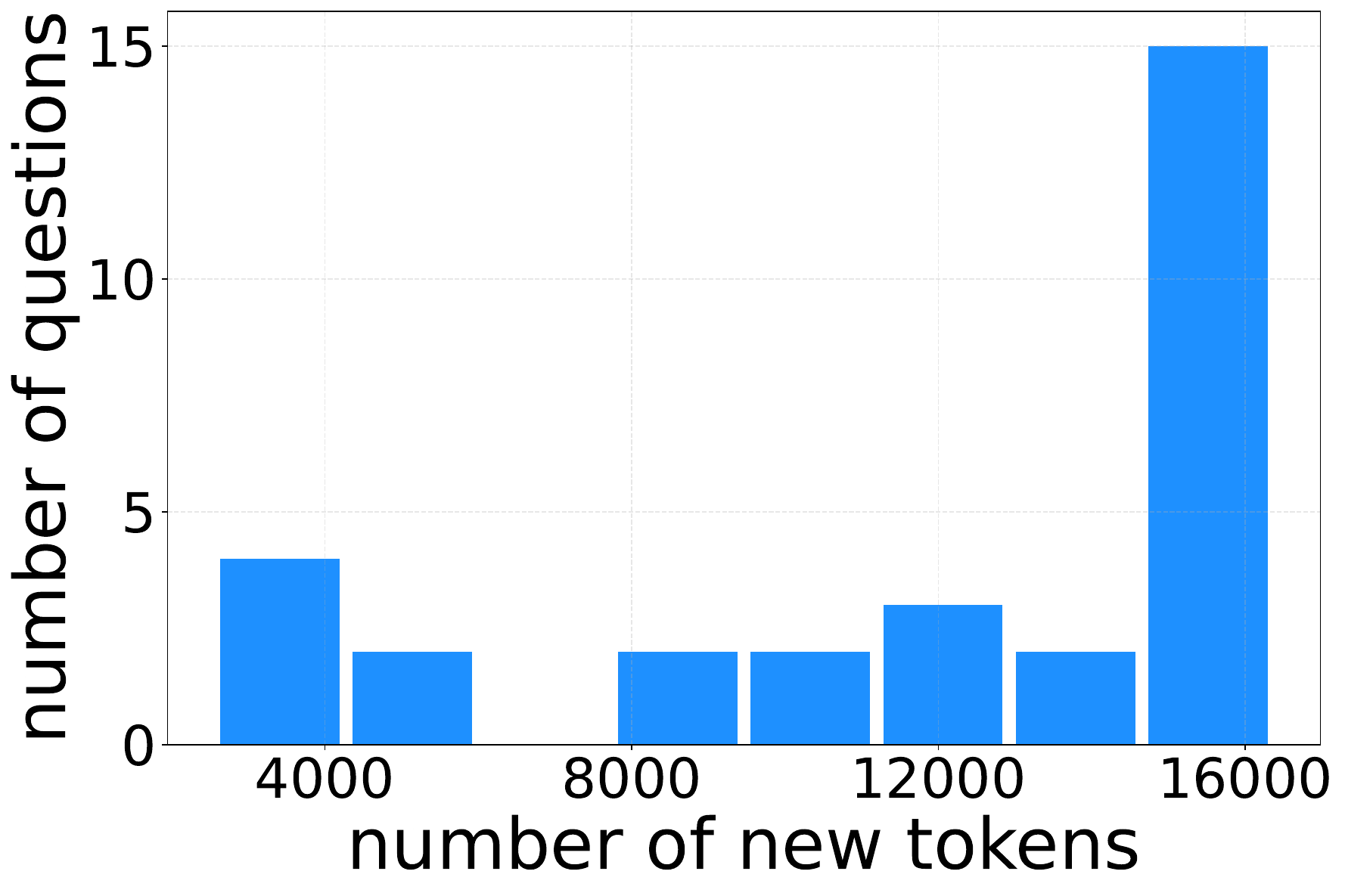}}
    \hfill
    \subfigure[\scriptsize\aime, correct vs. wrong \label{fig:aime_correct_wrong}]{\includegraphics[width=0.24\textwidth]{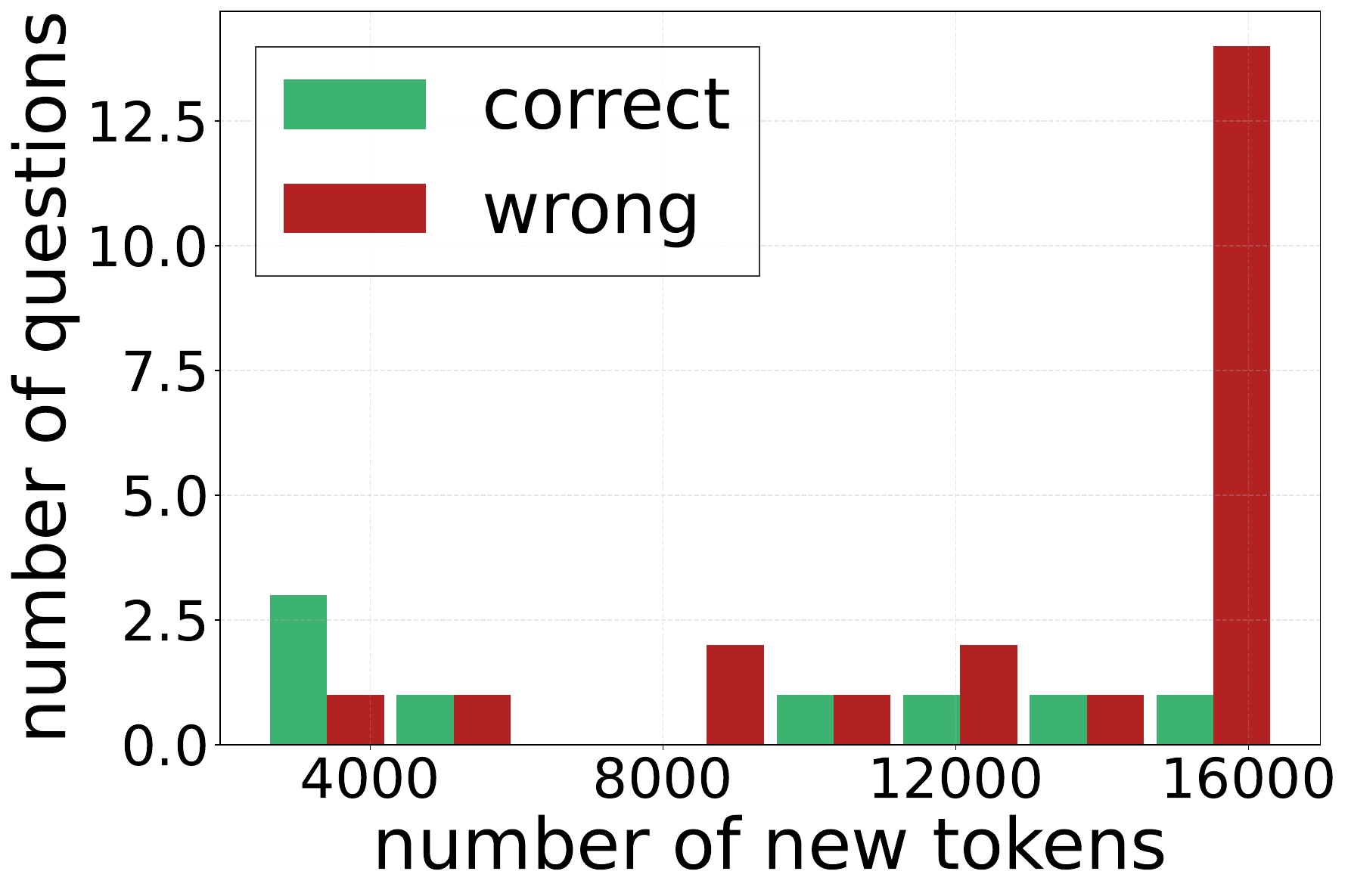}}
    \subfigure[\scriptsize \gsm, tail probability of correct answers \label{fig:gsm_tail_probability_correct}]{
  \includegraphics[width=0.23\textwidth]{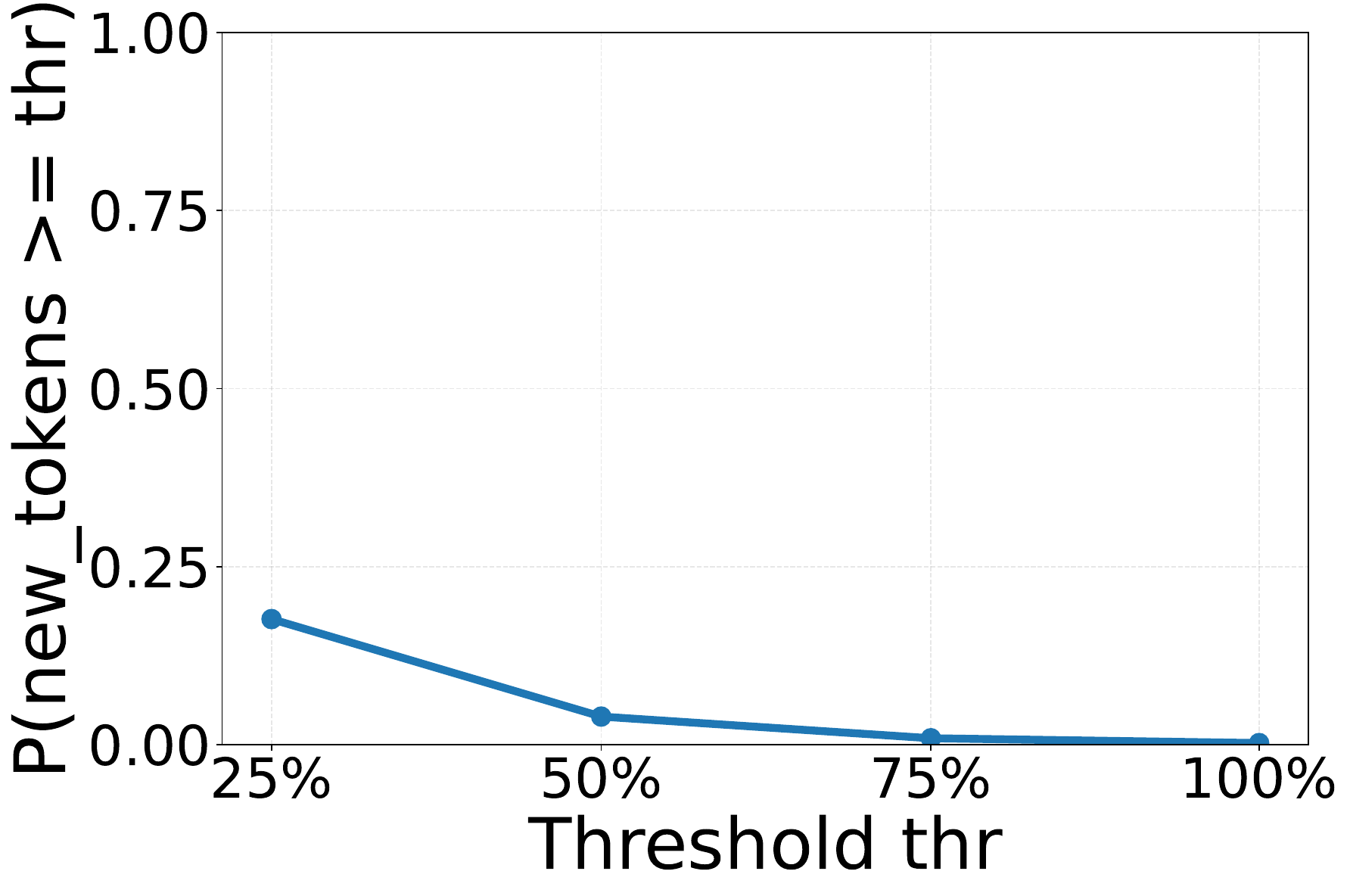}
    }
    \hfill
    \subfigure[\scriptsize \gsm, tail probability of wrong answers \label{fig:gsm_tail_probability_wrong}]{
  \includegraphics[width=0.23\textwidth]{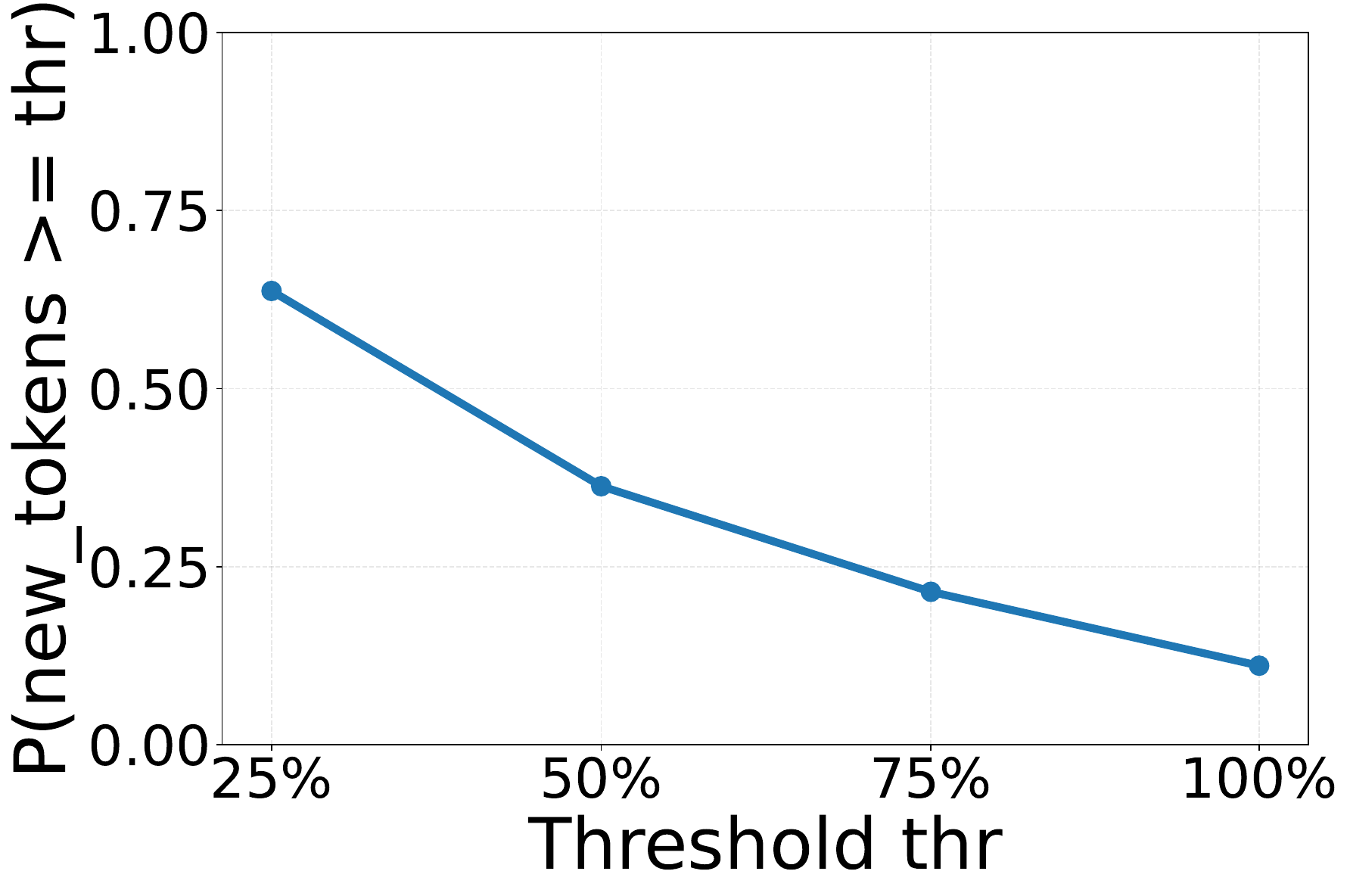}
    }
    \hfill
    \subfigure[\scriptsize \aime, tail probability of correct answers\label{fig:aime_tail_probability_correct}]{ \includegraphics[width=0.23\textwidth]{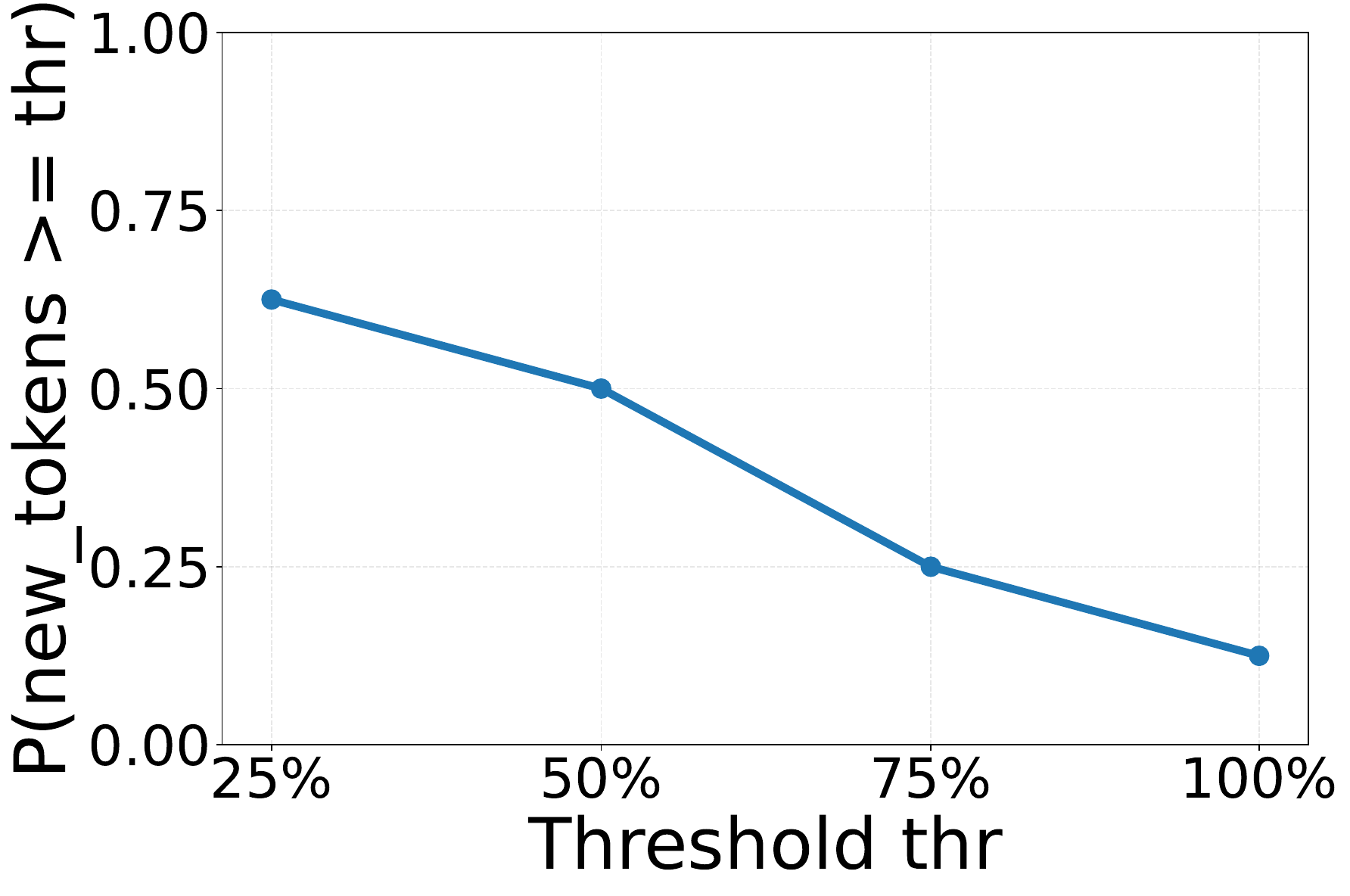}
    }
    \subfigure[\scriptsize \aime, tail probability of wrong answers \label{fig:aime_tail_probability_wrong}]{     \includegraphics[width=0.23\textwidth]{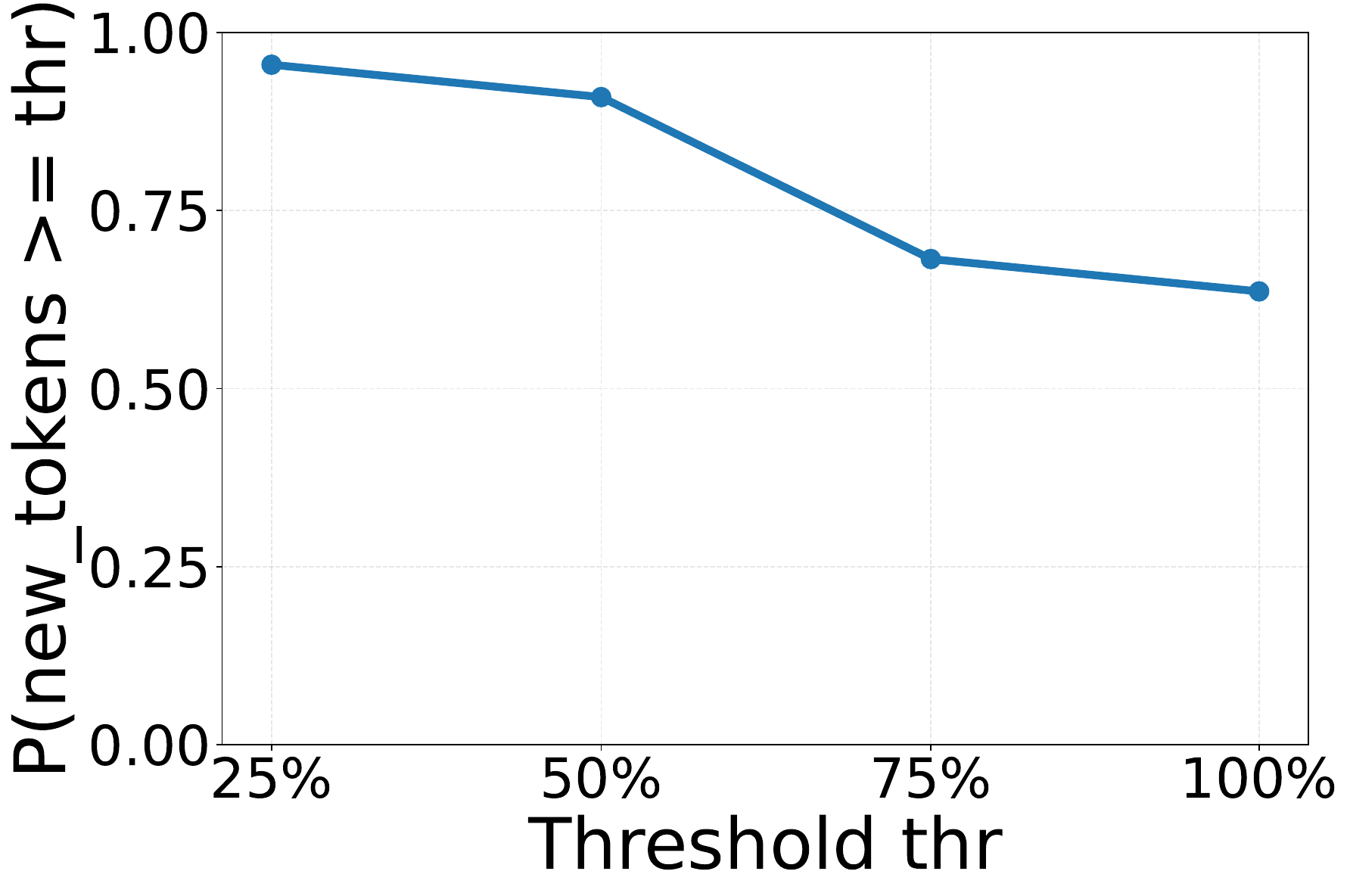}
    }
    \hfill
    \vspace{-2ex}
    \caption{Observation 2\,-\,Long generations are more likely to produce incorrect answers.}
    \label{fig:observation}
\end{figure*}

\paragraph{Observation 1: Attention approximation errors are more harmful at uncertain decoding steps.}
\rocketkv approximates attention scores by grouping keys into pages and using a subset of query dimensions to select top-$k$ keys~\cite{rocketkv2025}. At each decoding step, the model produces a probability distribution over possible next tokens. We consider a step uncertain when the probability is spread across multiple candidates, and more confident when it is concentrated on a few. We investigate whether uncertain steps exhibit larger key-selection errors and whether similar errors cause greater changes in the next-token distribution at uncertain decoding steps. 

We analyze DeepSeek-R1-Distill-Llama-8B on \aime with a KV cache token budget of 128, measuring uncertainty using the normalized entropy of the next-token distribution at the sampling temperature of 0.6. We group decoding steps into ten uncertainty deciles, from the most certain (decile 1) to the most uncertain (decile 10). Comparing \rocketkv's selected keys with exact top-$k$ selection at each step and layer, we find that \rocketkv preserves about 60\% of the attention mass retained by exact selection (58.87\%--62.39\%) and achieves around 19\% key overlap (18.13\%--20.38\%) as shown in Figures~\ref{fig:attention_vs_uncertainty} and~\ref{fig:overlap_vs_uncertainty}. Moreover, these errors remain nearly constant across uncertainty levels.

However, a different pattern emerges when examining how these errors affect model predictions. To isolate this effect, we perform an additional forward pass at each decoding step using the same KV cache but replacing \rocketkv's approximate key selection with exact top-$k$ selection. We measure the total variation (TV) distance between the resulting next-token distributions 
and flip rate, defined as the fraction of steps where the most likely token differs between the two distributions.

As shown in Figures~\ref{fig:total_variation} and~\ref{fig:flip_rate}, TV distance increases from 0.06 at the most certain steps to 0.47 at the most uncertain steps, an approximately $8\times$ increase, indicating that approximation errors substantially distort the sampling distribution when the model is uncertain. Similarly, the flip rate increases from 1.4\% to 65\%. To study the impact, we use Spearman correlation, which measures monotonic relationships using ranks, meaning it captures whether two variables generally increase or decrease together, even when the relationship is nonlinear. We use Spearman because we want to show that higher uncertainty generally corresponds to larger TV distance, without assuming a linear relationship. Across decoding steps from all 30 \aime problems, higher uncertainty generally corresponds to larger TV distance (i.e., Spearman’s rank correlation 0.65). To examine whether this relationship depends on the progress through a reasoning trace, we group decoding steps into ten bins based on their relative positions within each generated answer. 
The correlation stays almost the same (0.646), suggesting that the relationship holds across reasoning stages and is not due only to position in the trace. 

These results show that \emph{the impact of approximation errors of \rocketkv's algorithm depends more on decoding uncertainty than on error magnitude}. At confident steps, few tokens dominate the distribution, making predictions robust to approximation. At uncertain steps, competing tokens have similar probabilities, so the same error can change the selected token and potentially redirect subsequent reasoning. Indeed, replacing \rocketkv's approximate selection with exact top-$k$ improves \aime accuracy from 0\% to 16.67\% at the same budget. This motivates uncertainty-guided pruning: retain more keys at uncertain steps while pruning aggressively at confident ones. Uncertainty can be estimated from the preceding step's output distribution without an additional forward pass.

\vspace{-1ex}
\paragraph{Observation 2: Long generations are more likely to produce incorrect answers.}
Recent studies~\cite{rkv2025, skipkv2026} show that reasoning models generate many redundant and non-executive CoT tokens, creating opportunities for KV cache compression. However, identifying such tokens must be efficient, as the overhead of existing compression methods can offset their latency and throughput benefits~\cite{rethinkingserving2025, pagedeviction2026}. We analyze generation lengths on \gsm and \aime using DeepSeek-R1-Distill-Llama-8B, with $max\_new\_token$ values of 8192 and 16384, respectively. As Figures~\ref{fig:gsm_all_answers}--\ref{fig:aime_correct_wrong} show, most generations finish early, while a small fraction produce long traces, often corresponding to incorrect answers. We further measure the tail probability, i.e., the probability that generation length exceeds a threshold ranging from 25\% to 100\% of the decoding cap. Figures~\ref{fig:gsm_tail_probability_correct}--\ref{fig:aime_tail_probability_wrong} show that tail probabilities decrease rapidly for correct answers but remain high for incorrect ones. This suggests that prolonged generation often reflects repetitive reasoning or similar tokens rather than meaningful ones. This motivates our low-overhead redundancy-aware token scoring method, which aims to identify redundant tokens with high precision while preserving inference efficiency.

\vspace{-1ex}
\subsection{Uncertainty-guided Adaptive Pruning} 
Motivated by Observation 1, we dynamically adjust the pruning strength based on the model's decoding uncertainty.

\textbf{Uncertainty estimation.}
We use a standard uncertainty measure based on the normalized entropy of the next-token sampling distribution. Let $\mathbf{p}_t$ denote the predicted (sampling) distribution at step $t$, and let $H(\mathbf{p}_t)$ denote its entropy. The uncertainty score is defined as:
\begin{equation}
\label{eqn:unc-score}
u_t = \left(\frac{H(\mathbf{p}_t)}{\log |\mathcal{V}|}\right)^{\gamma},
\end{equation}
where $|\mathcal{V}|$ denotes the model's vocabulary size and $\gamma \in (0,1]$ is a monotonic transform that adjusts the dynamic range. By construction, $u_t \in [0,1]$, and larger values indicate greater uncertainty.

To avoid abrupt changes across neighboring steps, we smooth this signal using a moving average:
\begin{equation}
\label{eqn:smooth-avg}
\bar{u}_t = \beta \bar{u}_{t-1} + (1-\beta) u_t, \, t>1,
\end{equation}
where $\bar{u}_{t-1}$ is the smoothed uncertainty score at step $t-1$ and $\beta \in [0,1)$ controls the temporal smoothing. 

\textbf{Adaptive pruning control.}
We convert the smoothed uncertainty into a continuous relaxation factor using two thresholds $\tau_{\text{low}}$ and $\tau_{\text{high}}$:
\begin{equation*}
\alpha_t = \min\!\left(1,\ \max\!\left(0,\ \frac{\bar{u}_t - \tau_{\text{low}}}{\tau_{\text{high}} - \tau_{\text{low}}}\right)\right).
\end{equation*}
Here, $\alpha_t \in [0,1]$ controls the strength of pruning. When $\bar{u}_t$ is low (confident phase), $\alpha_t \approx 0$ and stronger pruning is applied. When $\bar{u}_t$ is high (uncertain phase), $\alpha_t \approx 1$ and the pruning is relaxed. Otherwise, the pruning strength is between relaxed and hard pruning. Note that the ${\tau_{\text{high}}}$ and ${\tau_{\text{low}}}$ are hyperparameters.

\begin{figure*}[!ht]
    \centering
\includegraphics[width=\textwidth]{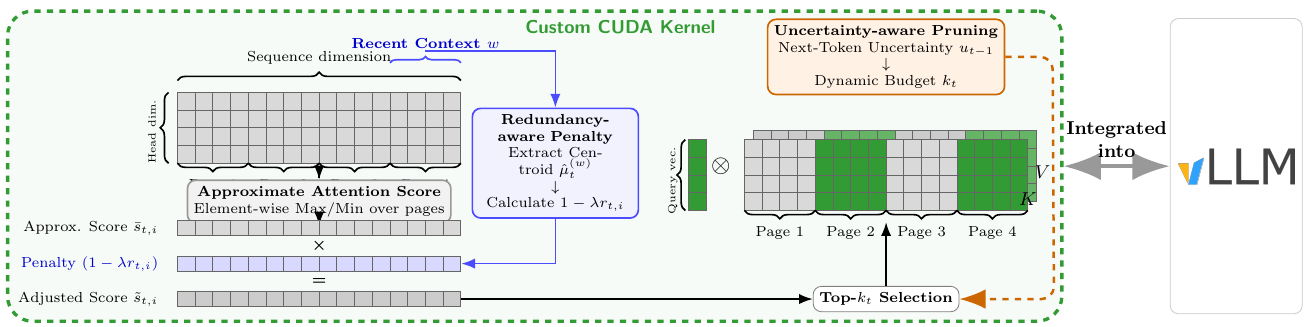}
    \vspace{-4ex}
    \caption{Overview of the \rorekv sparse attention pipeline.}
    \label{fig:rorekv-overview}
\vspace{-1ex}
\end{figure*}

\textbf{Integration with \rocketkv.}
\rocketkv~\cite{rocketkv2025} removes coarse-grain KV tokens with low importance after prefill. During the decode, as Figure~\ref{fig:rorekv-overview} demonstrates, it first groups cached key tokens into pages  and stores per-page maximum and minimum key values along the head dimension. These max/min values are updated while new tokens are being generated.
For each query, \rocketkv keeps only the most important query dimensions and uses the page max/min to approximate the maximum attention score of each page. It then selects the top-ranked pages, retrieves their original key and value vectors, and performs exact attention only over this smaller subset. \rorekv applies an uncertainty-guided approach on top of \rocketkv as illustrated in Figure~\ref{fig:rorekv-overview}.

Let $k^{\text{base}}$ and $\rho^{\text{base}}$ denote the top-$k$ budget and compression ratio used in \rocketkv. We adapt these parameters at each step by interpolating between a hard/relaxed pruning regime.
\begin{equation*}
\begin{aligned}
k_t & = (1-\alpha_{t-1}) \, k_{\text{hard}} + \alpha_{t-1} \, k_{\text{relax}},
\\
\rho_t & = (1-\alpha_{t-1})\, \rho_{\text{hard}} + \alpha_{t-1} \, \rho_{\text{relax}},
\end{aligned}
\end{equation*}
where $k_{\text{hard}} < k^{\text{base}} < k_{\text{relax}}$ and $\rho_{\text{relax}} < \rho^{\text{base}} < \rho_{\text{hard}}$. 
In this formulation, low uncertainty reduces the effective top-$k$ and increases pruning, while high uncertainty expands the set of tokens selected for attention. This design dynamically adjusts token selection during decoding, reducing unnecessary computation and memory bandwidth while preserving KV cache context, leading to improved computation and KV cache access efficiency.

\vspace{-2ex}

\subsection{Redundancy-aware Token Scoring}

Reasoning traces often exhibit redundancy, where multiple tokens convey highly similar information, leading to inefficient allocation of the KV cache budget. To address this, \rorekv introduces a redundancy-aware scoring mechanism that down-weights tokens that are semantically similar to the recent context.

At decoding step $t$, let $N_t$ denote the number of cached tokens, indexed by $i \in \{1,\ldots,N_t\}$. Let $\mathbf{k}_i \in \mathbb{R}^d$ denote the key vector of cached token $i$, and define its normalized version as $\hat{\mathbf{k}}_i = \mathbf{k}_i / \|\mathbf{k}_i\|_2$. We define $w$ as the redundancy window length, a hyperparameter controlling the number of recent cached keys used to construct the reference vector. The effective window length is $w_t=\min(w,N_t)$, and the reference vector is computed as:
\begin{equation}\label{centroid-eq}
\hat{\boldsymbol{\mu}}^{(w)}_t =
\frac{\sum_{j=N_t-w_t+1}^{N_t} \hat{\mathbf{k}}_j}
{\left\|\sum_{j=N_t-w_t+1}^{N_t} \hat{\mathbf{k}}_j\right\|_2}.
\end{equation}
The redundancy score for token $i$ is then defined using cosine similarity, denoted as $cosim$:
\begin{equation}\label{redundancy-eq}
r_{t,i} =
\begin{cases}
0, & i=N_t, \\
\max\!\left(0,\ cosim\!\left(\hat{\mathbf{k}}_i,\hat{\boldsymbol{\mu}}^{(w)}_t\right)\right), & 1 \le i < N_t 
\end{cases}
.
\end{equation}
Higher values of $r_{t,i}$ indicate that token $i$ is more aligned with the recent context and therefore more likely to be redundant. The condition $r_{t,N_t}=0$ excludes the most recent cached token from the redundancy penalty, preserving immediate local context.

Let $\bar{s}_{t,i}$ denote the non-negative approximate attention score used by \rocketkv for selecting cached token $i$ at decoding step $t$, prior to top-$k$ selection. We incorporate redundancy by applying a multiplicative adjustment to this selection score:
\begin{equation}\label{s-tild}
\tilde{s}_{t,i} = \bar{s}_{t,i} \left(1 - \lambda r_{t,i}\right),
\end{equation}
where $\lambda \in [0,1]$ controls the strength of the penalty. This formulation biases selection toward tokens that provide novel information, while suppressing repetitive tokens. Since $\bar{s}_{t,i}\ge 0$ and $(1-\lambda r_{t,i})\in[0,1]$, this operation can only decrease or preserve the selection score of redundant tokens (see Figure~\ref{fig:rorekv-overview}).

\textbf{Discussion.}
While \rorekv and other methods such as \rkv~\cite{rkv2025} aim to reduce redundancy, they differ in how redundancy is computed and integrated. \rorekv defines redundancy locally using a recent-context centroid, as shown in Equations~\ref{centroid-eq}--\ref{redundancy-eq}, requiring a single cosine similarity per token. In contrast, \rkv computes redundancy using global pairwise similarities across the KV cache. Specifically, let
\begin{equation}
\label{eqn:rkv-avg}
a^{\text{\rkv}}_{t,i}
=
\frac{1}{N_t-1}
\sum_{\substack{j=1 \\ j\ne i}}^{N_t}
 cosim\!\left(\hat{\mathbf{k}}_i,\hat{\mathbf{k}}_j\right), 
\end{equation}
denote the average similarity of token $i$ to all other cached tokens. \rkv then normalizes these scores across the cache:
\begin{equation}
\label{eqn:rkv-red}
r_{t,i}^{\text{\rkv}}
=
\left[
\mathrm{softmax}
\left(
\left[
a^{\text{\rkv}}_{t,1},
\ldots,
a^{\text{\rkv}}_{t,N_t}
\right]
\right)
\right]_i .
\end{equation}
It then combines this redundancy score additively with attention-based importance. This requires computing and aggregating token-to-token similarities over the entire cache. Compared to \rkv, \rorekv provides a lightweight and scalable approximation to redundancy by leveraging a compact summary of the recent context. This design avoids the high cost of pairwise similarity computation, reduces memory overhead, and aligns naturally with the temporal structure of decoding. Empirically, this allows \rorekv to suppress repetitive reasoning patterns while maintaining high fidelity, leading to more efficient KV utilization and improved accuracy under tight token budgets.

Beyond our primary token-level pruning approach, we further investigated a broader spectrum of pruning strategies, including layer-wise pruning, thought (sentence)-level pruning, and hybrid pruning that combines token- and thought-level granularity. These additional explorations highlight the breadth of our investigation and provide valuable insights into the potential of pruning at different levels of granularity. We present these extensions and their corresponding analyses in the Appendix.

\subsection{\vllm Integration}
\label{sec:vllm-integration}

We integrate \rorekv into \vllm without modifying any \vllm internals. The
integration attaches through \vllm's standard plugin mechanism
(\texttt{platform\_plugins}) and registers a custom attention backend
that handles decode steps, where KV cache traffic dominates runtime for
long reasoning traces. 
This preserves \vllm's scheduler, allocator,
and block manager, and ensures that paged memory management, prefix
caching, and CUDA graph capture continue to work without any change to the underlying \vllm source code.

\textbf{Selection, not reformatting.}
\rorekv accelerates attention by selecting which cached tokens to
attend to at each decoding step, without changing how cache entries are
stored. This is a key distinction from approaches such as
DeltaKV~\cite{deltakv2026}, which replace cached entries with a
compressed representation and therefore require an inline decompressor at
every attention step, together with matching changes to allocation and
prefix caching. Because \rorekv does not evict any pages and keeps the KV page table intact, our
kernels read directly from the KV cache using \vllm's native page table, which is what
makes the drop-in plugin integration above possible. 

\textbf{Three-phase decode.}
For our \vllm implementation, each decode step decomposes into three phases
that operate over the paged KV in a single end-to-end pipeline. We implement the redundancy-aware scoring and the uncertainty-guided adaptive  pruning of
Section~\ref{sec:3-proposed} in a way that avoids extra passes over the
KV cache and any CPU coordination.

\paragraph{Phase A: approximate scoring and redundancy.}
A streaming kernel walks the KV cache once. For each cached token $i$, it
loads the key vector $\mathbf{k}_i$ from paged memory, computes the
cosine similarity to the recent-context summary
$\hat{\boldsymbol{\mu}}^{(w)}_t$ (the redundancy score $r_{t,i}$ defined in
Equation~\eqref{redundancy-eq}) and per-token logits. Note that we compute scores for every single token. What separates this from the preliminary pass of an exact top-k attention algorithm is the strategies we take to reduce memory bandwidth. Since decode is primarily memory-bound, reducing memory traffic can greatly improve performance. To do this, we experimented with two strategies for cutting memory bandwidth which centered on loading only certain hidden dimensions of the key cache.
vLLM's default key cache layout is \texttt{[num\_pages, num\_tokens\_per\_page, num\_kv\_heads, head\_dim]}, which we refer to as NHD. \rocketkv{} and \rorekv{} select the top-$r$ entries of the query along the head dimension and then perform sparse loads from the key cache at those indices. 

Under NHD, these sparse loads index the innermost, contiguous dimension. Because GPUs fetch memory in fixed-size contiguous segments regardless of how many bytes within a segment are actually requested, the sparse loads end up fetching nearly the entire key cache, negating the bandwidth savings of sparsity. To address this, we introduce an alternative layout, which we call key-transposed (KT), in which the key cache is stored as \texttt{[num\_pages, num\_kv\_heads, head\_dim, num\_tokens\_per\_page]}. 

The only drawback of the KT layout is when we finalize our token selection and wish to do attention with them, we may have sparse loads. In this case, the NHD layout is more efficient.  
This is especially a mitigating factor for performance since the top-k kernel may produce tokens in an order such that no tokens in a contiguous block are in the same page. To coalesce accesses in KT, we sort the list of tokens selected for attention by token index such that tokens in the same page can be loaded in the same accesses. This phase is not present in the NHD kernels, as it is not a necessary optimization there. We experimented with both NHD and KT kernels and found that KT was better suited towards longer sequences, where the number of tokens needed to be processed by phase A was significantly larger than our actual token budget and therefore reducing the memory traffic of streaming the KV cache became the dominant factor.

\textbf{Softmax and redundancy injection.}
A small reduction kernel converts the per-page logits into approximate
softmax scores $\bar{s}_{t,i}$, aggregates these scores across
heads within each GQA group,   and applies the redundancy multiplier from
Equation~\eqref{s-tild} to obtain the adjusted scores
$\tilde{s}_{t,i} = \bar{s}_{t,i}\,(1 - \lambda r_{t,i})$ in a single
pass. The redundancy term reuses the per-position cosine similarities
that Phase A already wrote, so no additional read of the KV cache is
needed.

\paragraph{Phase B: top-$k$ selection.}
Phase B selects, for every request and KV head, the $k_t$ cached tokens
with the largest adjusted scores $\tilde{s}_{t,i}$. It reads only the
fp32 score row written by the Phase A reduction, never the KV cache. We
use a custom CUDA kernel: one thread block per (request, KV head) pair
streams once through the row, each thread keeping its four best
candidates in registers, and a single block-wide radix sort then picks
the $k_t$ best of the block's candidates and writes them into a
persistent index buffer whose shape is fixed for the whole run. The
selection is approximate: a token of the exact top-$k_t$ is missed only
when more than four of the exact top-$k_t$ tokens fall into the same
thread's slice. All reported results use this kernel; an exact
\texttt{torch.topk} path is kept for validation. Under the KT layout, a
second small kernel re-sorts the selected indices by token position so
that Phase C's gathers from the transposed key cache are coalesced; the
selected set is unchanged. Appendix~\ref{app:topk} gives the details.

\paragraph{Phase C: exact sparse attention.}
Phase C computes attention only over the $k_t$ tokens chosen in Phase B.
Inspired by the online softmax of FlashAttention~\cite{flashattention2022},
the kernel streams the selected keys and values from the cache in small
groups (tiles), and incrementally updates the running attention output
and softmax statistics as each tile is processed. As a result, the full
attention matrix is never materialized, and the cost of Phase C scales
with $k_t$ rather than with the full cached sequence length.

\textbf{Adaptive budgets without CPU synchronization.}
Both the adaptive budget $k_t$ and the compression ratio $\rho_t$  vary
across decoding steps. A naive implementation would resolve them on the
CPU every step, breaking CUDA graph capture and stalling the pipeline.
We avoid this entirely: $k_t$ is delivered as a device tensor and read
inside the kernel, and the CPU never reads any tensor value back during
decoding. As a result, the entire three-phase pipeline, including the
per-request adaptive budgets, runs inside a single captured CUDA graph.

\textbf{Split-KV parallelism.}
Decode attention is strongly memory-bound and contributes only a single
query token per request per step, so the natural axes of parallelism are
the batch and head dimensions. When the batch is small relative to the
number of streaming multiprocessors, parallelizing only over $(B, H)$
leaves much of the device idle and fails to hide HBM latency, where $B$ is batch-size, and $H$ is number of heads. We
therefore use \emph{split-KV} execution: each request's KV cache is
partitioned into $S$ segments along the sequence dimension, each segment
is processed by a separate thread block to produce a partial attention
output, and a lightweight reduction kernel merges the partials. This
adds a third axis of parallelism $(B, H, S)$. We utilize a greedy work distribution algorithm that greedily assigns Cooperative Thread
Arrays (CTAs) to splits dynamically such that longer requests in our batch get more CTAs. 

Definition of some terminologies, implementation details,  including the kernel-level data layout, the
on-chip top-$k$ implementation, the split-KV work distribution, the
precise boundary handling in Phase C, and some tips for implementing custom attention algorithms into \vllm are deferred to
Appendix~\ref{app:integration}.

\vspace{-1ex}
\section{Experiments}
\label{sec:4-exp}
\subsection{Experimental Settings}
\rorekv\ is evaluated on two reasoning-oriented model families, DeepSeek-R1-Distill-Llama-8B and DeepSeek-R1-Distill-Qwen-7B~\cite{deepseek2025}, both based on GQA. We refer to these models as \deepllama and \deepqwen, respectively. Three mathematical reasoning benchmarks, \gsm\cite{gsm2021}, \mathf\cite{math2021} and \aime\cite{aime2025} are used as downstream tasks. \rorekv\ is compared against the \fullkv baseline and several KV cache compression baselines, including \rocketkv\cite{rocketkv2025}, \raas\cite{raas2025},  \rkv~\cite{rkv2025}, and SkipKV~\cite{skipkv2026}. We carefully implement all the baselines in our repository to be able to fairly evaluate and compare them with \rorekv. The maximum number of new tokens is set to 2048 for \gsm\ and to 16384 for \aime\ and \mathf. Experiments are conducted across a range of KV token budgets for each dataset and model. For \rorekv, the KV token budget reflects the average memory traffic used for \topk in the attention module. We set the adaptive pruning parameters relative to the base configuration as $k_{\text{hard}} = 0.9k^{\text{base}}$, $k_{\text{relax}} = 1.3k^{\text{base}}$ (with $k^{\text{base}} \in \{32, 64, 128, 256\}$), $\rho_{\text{hard}} = 1.5\rho^{\text{base}}$, and $\rho_{\text{relax}} = 0.9\rho^{\text{base}}$ with $\rho^{\text{base}}=1$. Following other related work, including TriAttention~\cite{TriAttention2026}, the reported accuracy results in Section~\ref{sec:acc} are based on \passone~\cite{passk2021}, computed by averaging correctness over 8 generated responses per question for \aime and 1 generated response for \mathf and \gsm. For both models, we use a sampling temperature of 0.6 and a top-$p$ value of 0.95. We first compute the average \topk used in \rorekv and then run other baselines with that \topk value.  
For rotary position embeddings, the RoPE base is set to 500K for \deepllama\ and 10K for \deepqwen. 
The observation window size used in the first stage of \rocketkv is set to 32 for \gsm\ and \aime, and to 16 for \mathf. 
For \rorekv, the lower and upper uncertainty thresholds ($\tau_{\text{low}}$ and $\tau_{\text{high}}$), which govern uncertainty-guided adaptive pruning, are selected from the ranges [0.0, 0.2] and [0.5, 0.8], respectively. The redundancy penalty coefficient $\lambda$ is set to  0.1, and the redundancy window length $w$ is set to 8. 
The values of $\gamma$ and $\beta$ are  0.1 and 0.85 in Equations~\eqref{eqn:unc-score} and~\eqref{eqn:smooth-avg}, respectively. The experiments are run on A40 GPUs \footnote{We release the source code upon the acceptance of the paper.}.

\vspace{-1ex}
\subsection{Accuracy Results}
\label{sec:acc}

\begin{figure*}[!ht]
    \centering
  {\includegraphics[width=0.85\textwidth]{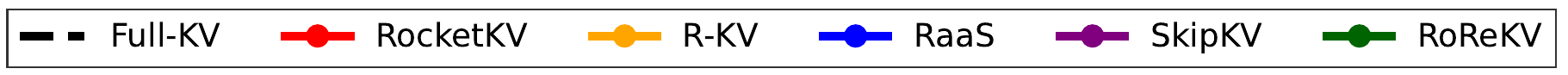}}
    \vspace{-1ex}
    \subfigure[\scriptsize \gsm, \deepllama \label{fig:acc_gsm_deepllama}]{  
    \includegraphics[width=0.31\textwidth]{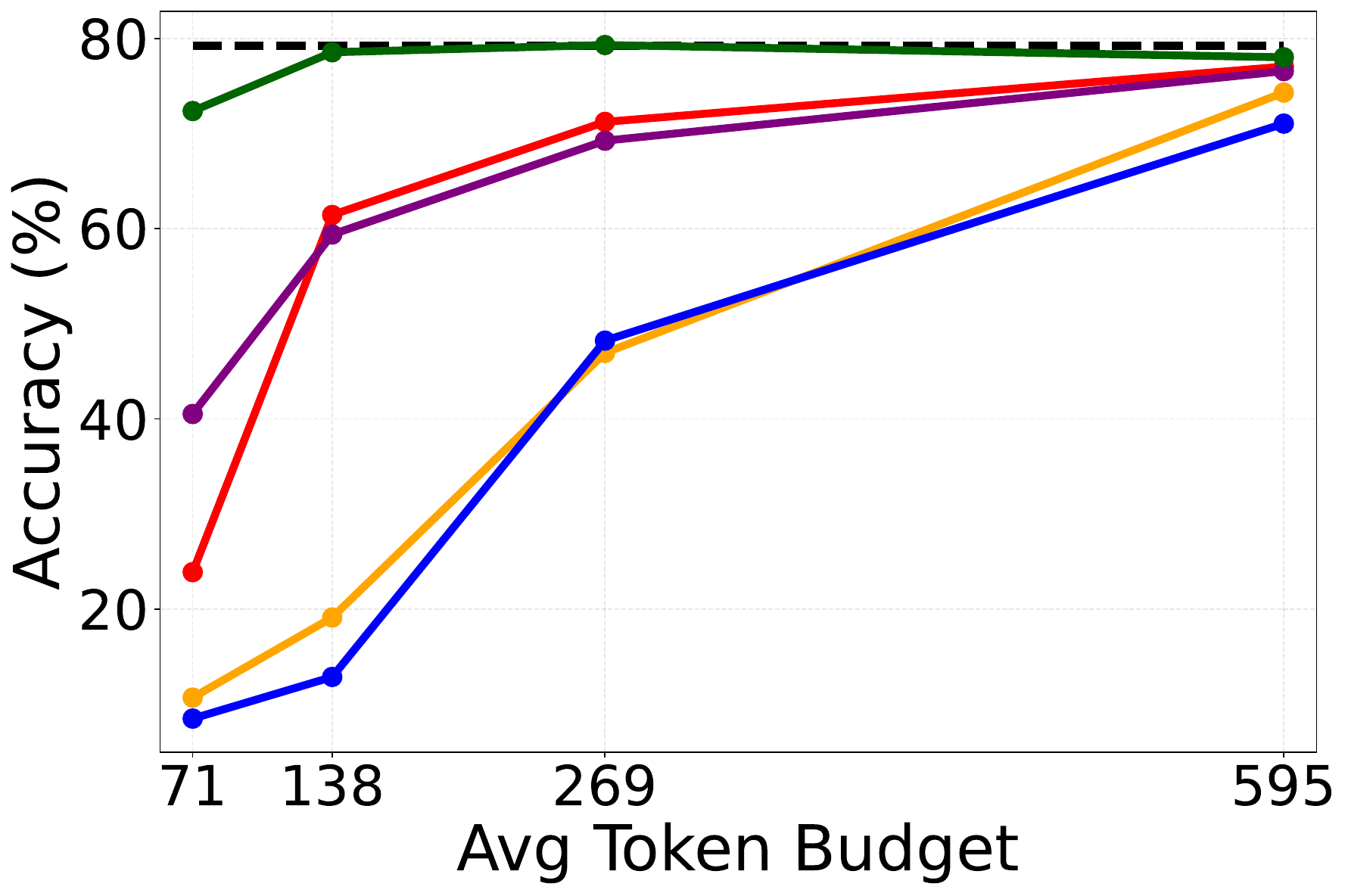}}
    \hfill
    \subfigure[\scriptsize \gsm, \deepqwen \label{fig:acc_gsm_deepqwen}]{%
    \includegraphics[width=0.31\textwidth]{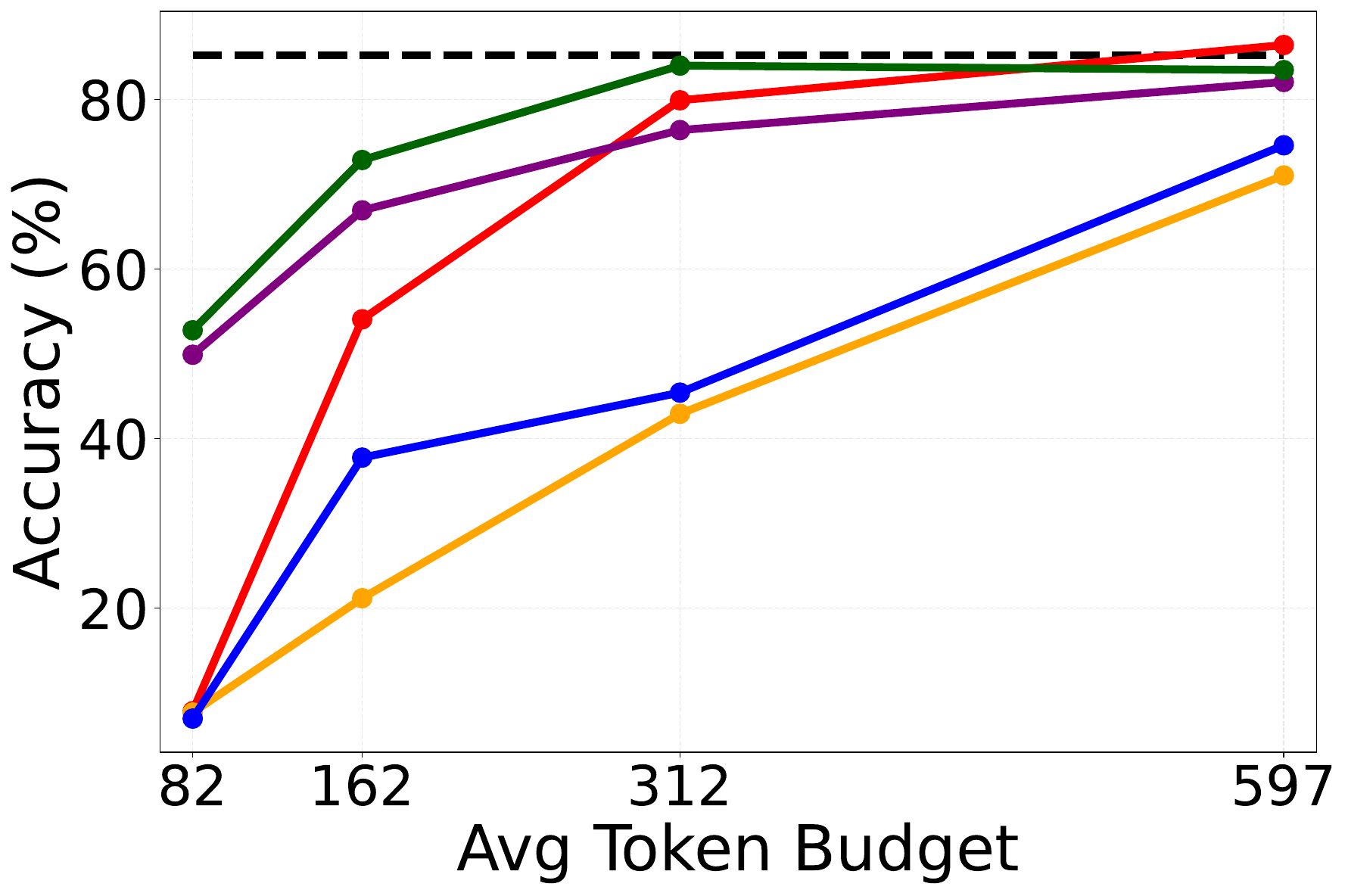}}
    \hfill
    \subfigure[\scriptsize \mathf, \deepllama \label{fig:acc_mathf_deepllama}]{\includegraphics[width=0.321\textwidth]{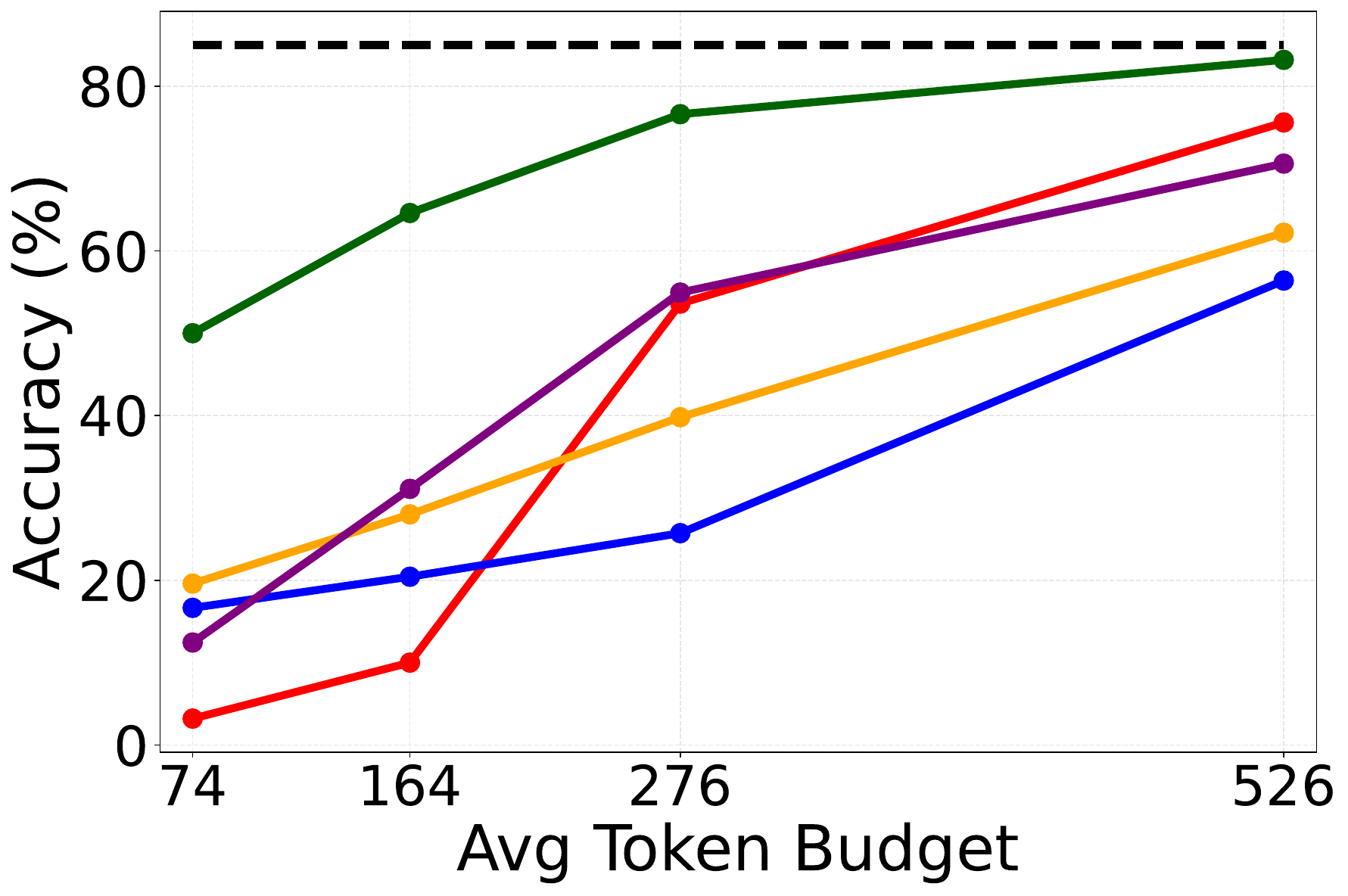}}
    \subfigure[\scriptsize \mathf, \deepqwen \label{fig:acc_mathf_deepqwen}]{
  \includegraphics[width=0.31\textwidth]{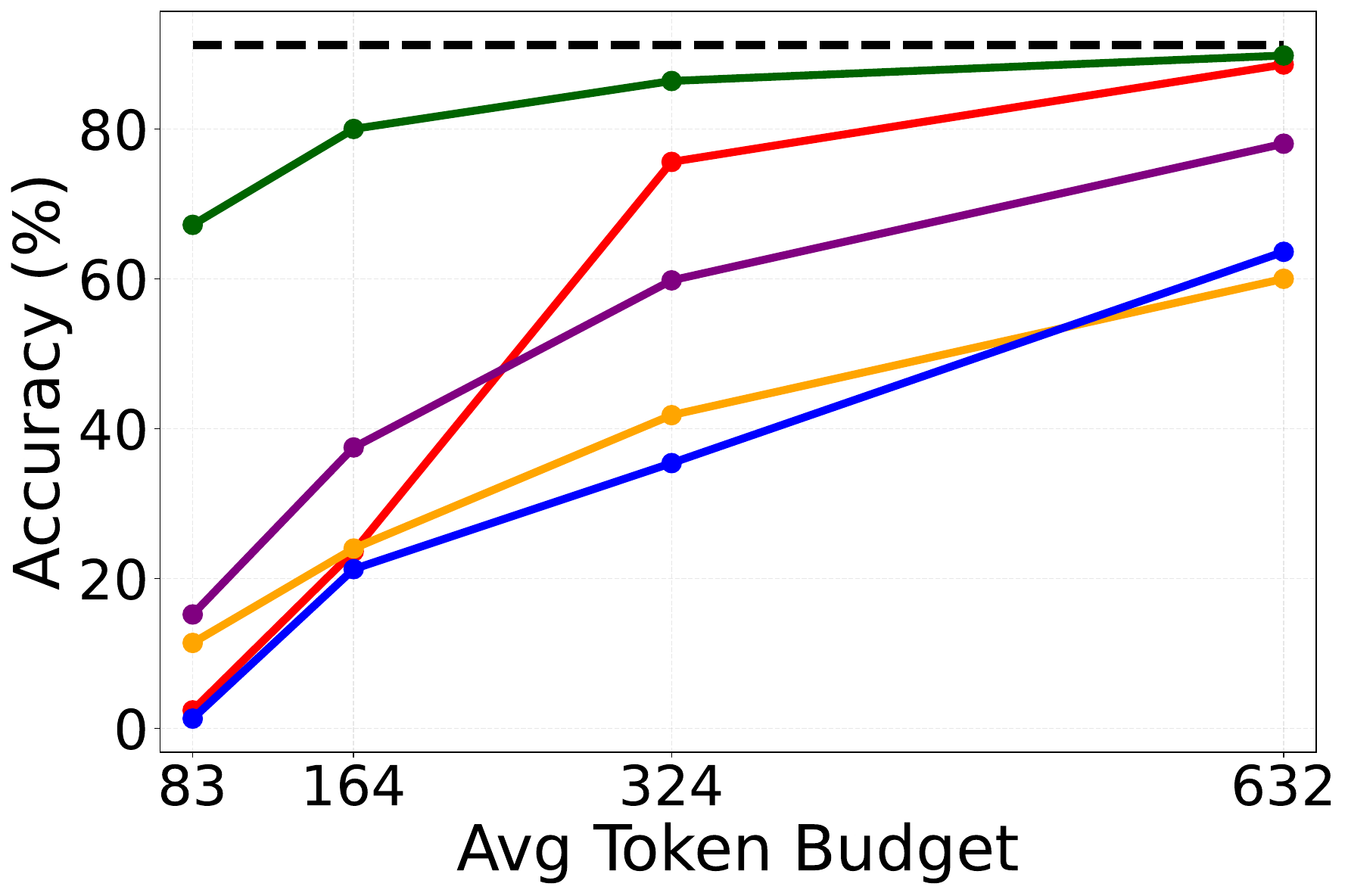}
    }
    \hfill
    \subfigure[\scriptsize \aime, \deepllama \label{fig:acc_aime_deepllama}]{
  \includegraphics[width=0.31\textwidth]{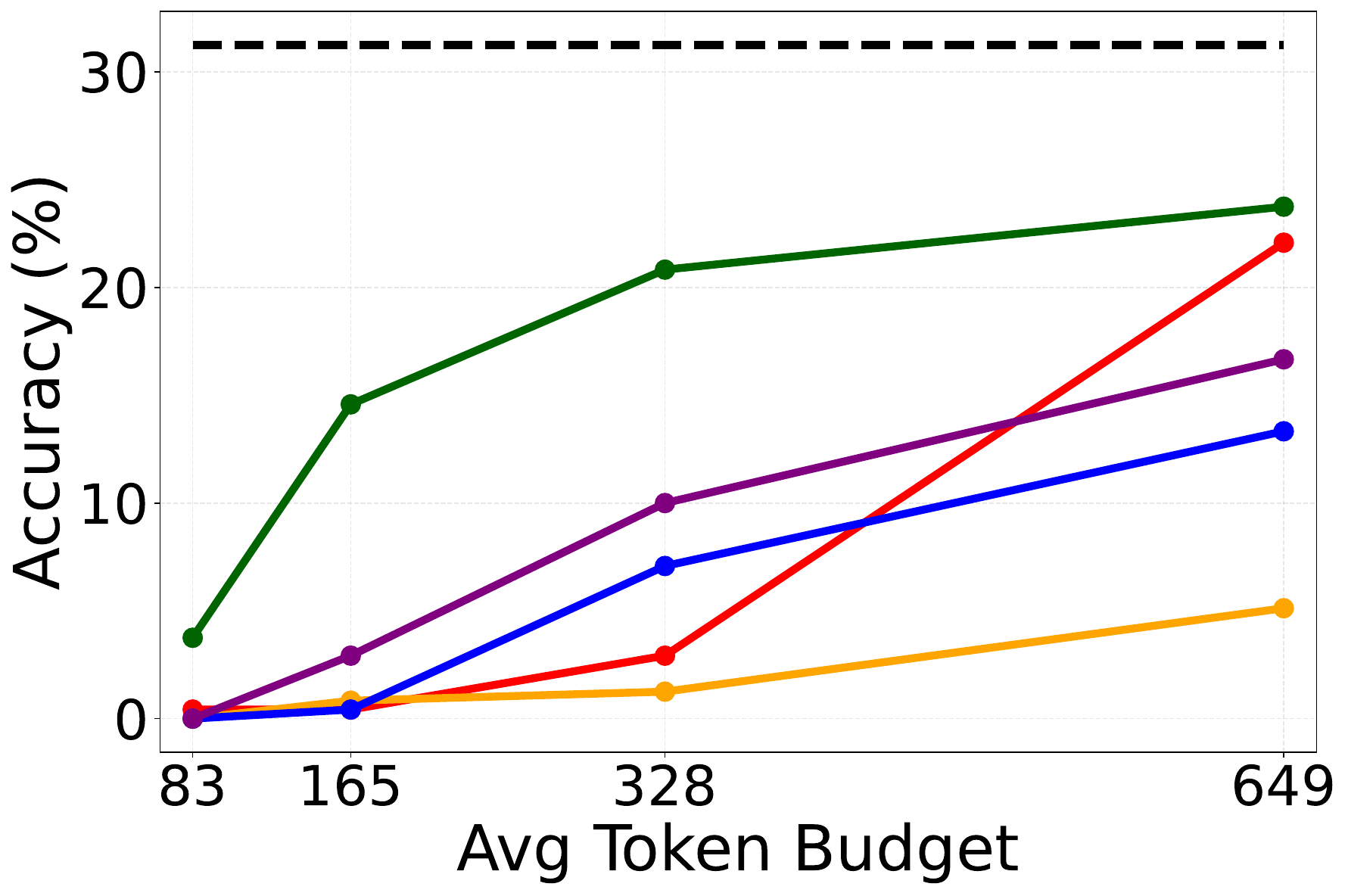}
    }
    \hfill
    \subfigure[\scriptsize \aime, \deepqwen \label{fig:acc_aime_deepqwen}]{     \includegraphics[width=0.31\textwidth]{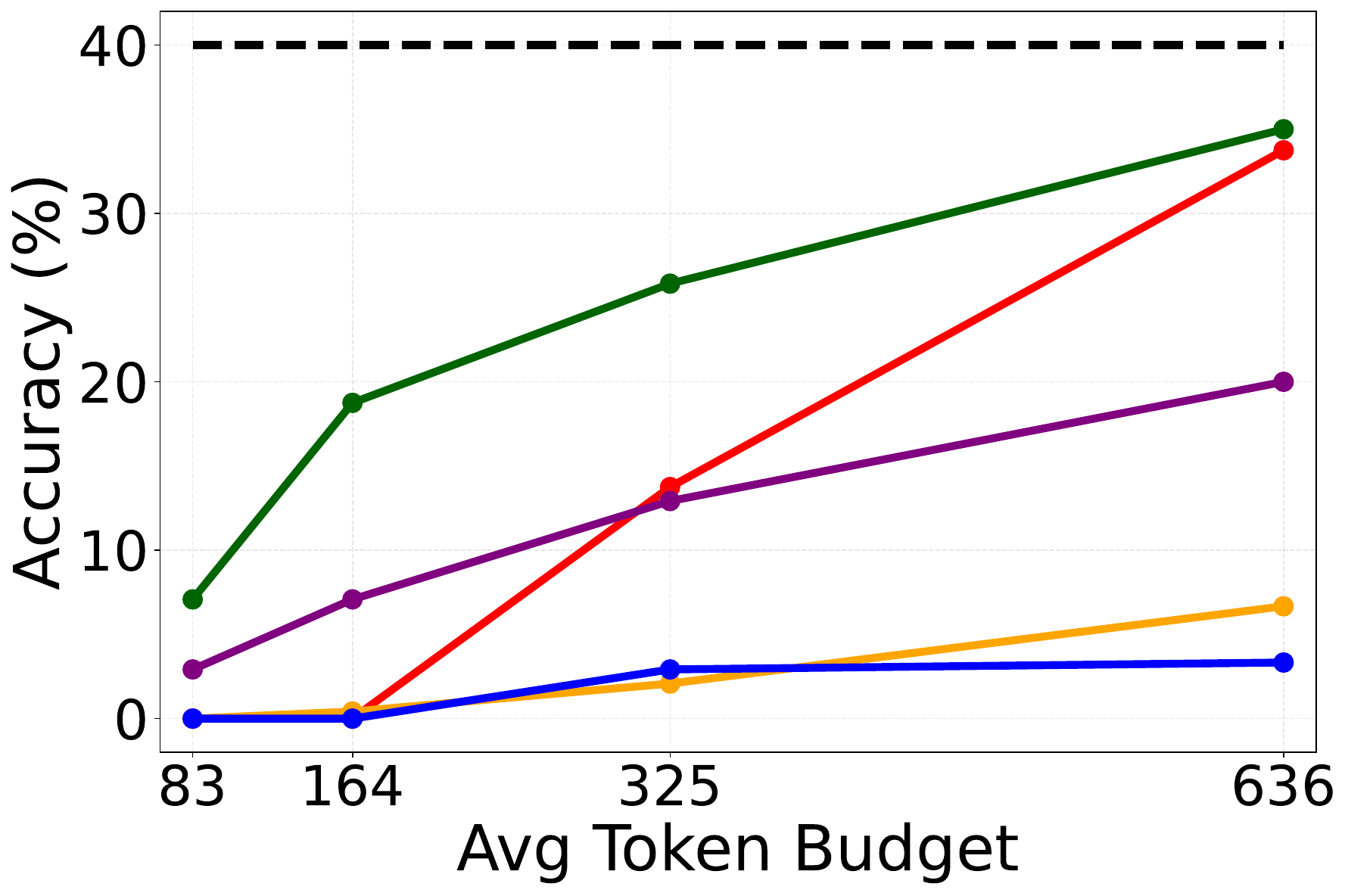}
    }
    \vspace{-2ex}
    \caption{Comparing the accuracy  of \rorekv with other baselines and \fullkv on \gsm  (a-b), \mathf (c-d), and \aime (e-f).}
    \label{fig:accuracy-model-dataset}
\end{figure*}

Figure~\ref{fig:accuracy-model-dataset} reports the accuracy of \rorekv\ compared to \fullkv\ and prior KV compression methods across different token budgets. Overall, \rorekv\ consistently achieves the best accuracy among all compression methods. 

On \gsm, \rorekv\ shows strong accuracy across both models. 
 Notably, a key distinguishing feature of \rorekv on \gsm is its ability to maintain relatively high accuracy even at the smallest token budget, whereas competing methods exhibit substantially lower accuracy under the same constraint. On \mathf, which requires longer reasoning, the advantage of \rorekv\ becomes more pronounced. At budget 74  on \deepllama and 83 on \deepqwen, \rorekv\ significantly outperforms \rocketkv\ by more than $+46\%$ on \deepllama  and $+65\%$ on \deepqwen. Even at larger budgets, \rorekv\ consistently remains closest to \fullkv. 

On the more challenging \aime\ benchmark, most compression methods suffer substantial accuracy degradation under tight KV budgets. In contrast, \rorekv demonstrates greater robustness, consistently outperforming the competing compression baselines. At averaged token budgets of 328 and 325, \rorekv\ achieves accuracies of 20.83\% and 25.83\% on \deepllama\ and \deepqwen, respectively, which are higher than other baselines. As the budget increases, \rorekv\ continues to outperform all competing methods, reaching 23.75\% on \deepllama\ and 35.00\% on \deepqwen on the largest evaluated budgets, narrowing the accuracy gap relative to \fullkv\ to just 7.5 and 5.0 percentage points, respectively.

These results show that \rorekv\ is effective at preserving critical reasoning information even under high compression compared to other state-of-the-art methods. By combining uncertainty-guided pruning with redundancy-aware token selection, it allocates the KV budget more efficiently, leading to robust accuracy across datasets, models, and budget regimes.

Compared with SkipKV, \rorekv\ achieves substantially higher accuracy, particularly under tight KV cache budgets. For example, on \mathf\ with \deepqwen, \rorekv\ improves accuracy from 15.20\% to 67.20\% at an average budget of 83 tokens, a gain of 52\%. Inspired by SkipKV, we also explored thought-level (sentence-level) pruning, which considers entire reasoning segments rather than individual tokens as the units of pruning and also hybrid-level pruning (thought-level combined with token-level uncertainty) built on \rocketkv backbone, with its design and experimental results presented in Appendix~\ref{app:thought}.

\vspace{-1ex}

\subsection{Efficiency Results}
\label{sec:efficiency-results}

We measure end-to-end latency and throughput of \rorekv on NVIDIA A40
GPUs running \deepllama on \aime and also synthetic benchmark data when the cache token budget is 512. We use a batch size of 16 for \aime and 6 for synthetic data. The synthetic benchmark allows us to evaluate the throughput of a sparse attention algorithm at longer context lengths. In our synthetic benchmark, we give the model a prompt and then sample tokens from the model until each request has reached a certain generation length. The reason we use synthetic data is to evaluate the efficiency of the proposed algorithm and \vllm integration on longer context lengths similar to other works~\cite{deltakv2026}.

\vspace{-1ex}
\begin{table}[h]
\centering
\caption{End-to-end efficiency on A40 over \deepllama. Latency and throughput are normalized to \fullkv.}
\label{tab:efficiency}
\vspace{-2ex}
\resizebox{0.5\textwidth}{!}{
\begin{tabular}{lcc}
\toprule
Method & Latency (s) $\downarrow$ & Throughput (tok/s) $\uparrow$ \\
\midrule
\fullkv (\vllm-\aime)   & 196.2 (1.00$\times$) & 247.3 (1.00$\times$) \\
\fullkv (\vllm-synthetic)  & 864.9 (1.00$\times$) & 115.0 (1.00$\times$) \\
\rkv~\cite{rkv2025} & 208.7 (0.94$\times$) & 276.9 (1.12$\times$) \\
\rorekv (\vllm-\aime)   & 154.5 (1.27$\times$) & 341.3 (1.38$\times$) \\
\rorekv (\vllm-synthetic)  & 604.8 (1.43$\times$) & 166.7 (1.45$\times$) \\
\bottomrule
\end{tabular}
}
\end{table}

Table~\ref{tab:efficiency} compares latency and throughput on A40 for the \vllm{} implementations of \fullkv, \rkv, and \rorekv. The \rorekv{} integration uses the fused three-phase kernels described in Section~\ref{sec:vllm-integration}.

Relative to \fullkv, \rkv achieves a latency speedup factor of 0.94$\times$, indicating a slowdown, and 1.12$\times$ throughput. On \aime{} and the synthetic benchmark, \rorekv{} achieves latency speedups of 1.27$\times$ and 1.43$\times$, respectively, with throughput reaching 1.38$\times$ and 1.45$\times$ the baseline.
So, \rorekv outperforms \rkv in both accuracy and efficient results.

\vspace{-2ex}
\subsection{Ablation Study}
\label{sec: ablation}
In this subsection, we conduct a series of ablation studies to further demonstrate the importance of each individual component on the final accuracy. Two basic ablations are first considered. The first one, denoted as \textbf{Unc.}, disables redundancy-aware token scoring and keeps only the uncertainty-guided adaptive pruning component. As shown in Table~\ref{tab:ablation_components}, the impact of this component on the final accuracy is higher than that of the redundancy-aware component.

The second basic one, denoted as \textbf{Redun.}, disables uncertainty-guided adaptive pruning and keeps only the redundancy-aware token scoring component. As the results show, this component by itself helps to reach a reasonably high accuracy (e.g., \deepllama on \gsm); however,  as Table~\ref{tab:ablation_components} illustrates, its role on the \aime dataset is lower compared to other datasets. Both uncertainty-guided and redundancy-aware components are complementary to each other, and in all experiments, the accuracy of \rorekv is higher than  the accuracy of each of these two components separately. 

As a third ablation, we study \textbf{Reverse}, which inverts the uncertainty-guided pruning decisions to test whether the direction of pruning is essential to its effectiveness. Reverse consistently underperforms \rorekv across all model-dataset pairs. For example, accuracy drops from 80.00\% to 5.80\% for \deepqwen on \mathf supporting the importance of the original pruning direction for maintaining reasoning accuracy.



\begin{table*}[!ht]
\centering
\caption{Component ablation of \rorekv.
Pass@1 accuracy (\%) at token budget 128.
\textbf{Redun.} and \textbf{Unc.} isolate
redundancy-aware scoring and uncertainty-guided
adaptive pruning, respectively.
\textbf{Reverse} inverts the pruning selection.}
\vspace{-2ex}
\label{tab:ablation_components}
\small
\setlength{\tabcolsep}{10pt}
\begin{tabular}{l c c c c}
\toprule
\textbf{Model / Dataset}
& \textbf{\rorekv}
& \textbf{Unc.}
& \textbf{Redun.}
& \textbf{Reverse} \\
\midrule
\deepllama / \aime
& \textbf{14.58} & 10.00 & 0.00 & 1.25 \\

\deepqwen / \aime
& \textbf{18.75} & 11.25 & 0.00 & 0.00 \\

\deepllama / \gsm
& \textbf{78.54} & 72.02 & 59.36 & 51.40 \\

\deepqwen / \gsm
& \textbf{72.86} & 68.92 & 33.66 & 31.01 \\

\deepllama / \mathf
& \textbf{64.60} & 57.40 & 4.60 & 3.40 \\

\deepqwen / \mathf
& \textbf{80.00} & 78.40 & 9.20 & 5.80 \\
\bottomrule
\end{tabular}
\end{table*}

We further investigate the sensitivity of \rorekv to its key hyperparameters, including the lower and upper uncertainty thresholds ($\tau_{\text{low}}$ and $\tau_{\text{high}}$), redundancy penalty coefficient ($\lambda$), and redundancy window length ($w$). As shown in Tables~\ref{tab:ablation_uncertainty} and~\ref{tab:ablation_redundancy}, \rorekv exhibits relatively stable accuracy across a range of hyperparameter settings, particularly for the redundancy window and lower uncertainty threshold. For example, varying $w$ from 4 to 15 results in less than a one percentage-point accuracy difference on \gsm for both models. Similarly, varying $\tau_{\text{low}}$ produces moderate accuracy fluctuations across the evaluated model--dataset combinations. These observations suggest that \rorekv does not require precise tuning of these hyperparameters to maintain its effectiveness.

\begin{table*}[!ht]
\centering
\caption{Ablation study on the lower ($\tau_{\text{low}}$)
and upper ($\tau_{\text{high}}$) uncertainty thresholds.
Results are pass@1 accuracy (\%) at a token budget of 128.}
\vspace{-2ex}
\label{tab:ablation_uncertainty}
\small
\setlength{\tabcolsep}{5pt}
\begin{tabular}{l c cccc cccc}
\toprule
\textbf{Model / Dataset}
& \textbf{\rorekv}
& \multicolumn{4}{c}{\textbf{Lower Threshold ($\tau_{\text{low}}$)}}
& \multicolumn{4}{c}{\textbf{Upper Threshold ($\tau_{\text{high}}$)}} \\
\cmidrule(lr){3-6}
\cmidrule(lr){7-10}
& & 0 & 0.05 & 0.1 & 0.2
  & 0.5 & 0.6 & 0.7 & 0.8 \\
\midrule
\deepllama / \aime
& \textbf{14.58}
& 6.67 & 9.58 & 9.58 & 7.92
& \textbf{14.58} & 10.42 & 6.67 & 12.08 \\

\deepqwen / \aime
& \textbf{18.75}
& 14.17 & 12.50 & 10.42 & 9.58
& \textbf{18.75} & 14.17 & 10.00 & 10.42 \\

\deepllama / \gsm
& \textbf{78.54}
& 71.57 & 72.18 & 69.98 & 72.86
& 70.74 & 71.95 & 71.57 & \textbf{78.54} \\

\deepqwen / \gsm
& \textbf{72.86}
& 67.85 & 69.90 & 70.43 & 69.52
& \textbf{72.86} & 70.43 & 66.49 & 60.12 \\

\deepllama / \mathf
& \textbf{64.60}
& 62.60 & 59.60 & 59.20 & 56.20
& \textbf{64.60} & 59.20 & 55.20 & 64.20 \\

\deepqwen / \mathf
& 80.00
& 80.00 & 79.60 & \textbf{80.20} & 79.40
& 72.80 & 65.20 & 61.20 & 59.60 \\
\bottomrule
\end{tabular}
\end{table*}
\vspace{-1ex}

\begin{table*}[!ht]
\centering
\caption{Ablation study on redundancy penalty coefficient ($\lambda$) 
and redundancy window length $w$.
Results are pass@1 accuracy (\%) at a token budget of 128.}
\vspace{-2ex}
\label{tab:ablation_redundancy}
\small
\setlength{\tabcolsep}{6pt}
\begin{tabular}{l c ccc ccc}
\toprule
\textbf{Model / Dataset}
& \textbf{\rorekv}
& \multicolumn{3}{c}{\textbf{Redundancy Penalty ($\lambda$)}}
& \multicolumn{3}{c}{\textbf{Redundancy Window ($w_t$)}} \\
\cmidrule(lr){3-5}
\cmidrule(lr){6-8}
& & 0.01 & 0.2 & 0.5 & 4 & 10 & 15 \\
\midrule
\deepllama / \aime
& \textbf{14.58} & 9.17 & 9.17 & 5.42
& 7.08 & 10.00 & 10.00 \\

\deepqwen / \aime
& \textbf{18.75} & 10.42 & 11.67 & 12.08
& 10.83 & 12.50 & 12.08 \\

\deepllama / \gsm
& \textbf{78.54} & 70.43 & 71.87 & 71.65
& 71.80 & 72.33 & 71.72 \\

\deepqwen / \gsm
& \textbf{72.86} & 66.34 & 72.71 & 76.12
& 68.54 & 68.01 & 68.54 \\

\deepllama / \mathf
& \textbf{64.60} & 60.60 & 59.40 & 58.40
& 59.20 & 59.20 & 59.20 \\

\deepqwen / \mathf
& 80.00 & 79.00 & 81.80 & \textbf{82.20}
& 80.40 & 79.00 & 77.80 \\
\bottomrule
\end{tabular}
\end{table*}
\vspace{-2ex}

\section{Related Work}

Recent work shows that KV cache compression for reasoning models is different from standard long-context compression, since most of the cost comes from long generated reasoning traces rather than the input prompt. Several methods, therefore, focus specifically on compressing the decode-stage of KV cache in large reasoning models~\cite{rkv2025,headkvr22025,TriAttention2026,rpc2025,skipkv2026}. 

HeadKV-R2~\cite{headkvr22025} allocates KV budgets at the head level by estimating each head's reasoning importance, showing that different heads contribute unequally during reasoning. 
RPC~\cite{rpc2025} compresses generated reasoning trajectories by exploiting semantic sparsity, while TriAttention~\cite{TriAttention2026} revisits token scoring for reasoning workloads and improves retention under long decoding budgets. More recently, Lethe~\cite{lethe2026} introduces layer-wise budget allocation and multi-round, recency-aware pruning during generation, improving throughput while maintaining reasoning quality. BeaconKV~\cite{beaconkv2026} selects a small set of diverse queries from previous decoding steps to identify KV tokens that may be important for future reasoning. It combines these queries with recent queries to score cached tokens, keeping the most important and recent tokens while removing less useful ones to reduce memory usage.
ForesightKV~\cite{foresightkv2026} learns future-aware KV eviction policies through supervised and reinforcement learning, improving reasoning accuracy under constrained cache budgets. 

In contrast, \rorekv extends a training-free KV compression pipeline and injects reasoning-aware signals directly into decode-time token selection. Specifically, \rorekv combines redundancy-aware token selection with uncertainty-guided adaptive pruning: the former down-weights repetitive tokens, while the latter relaxes or strengthens pruning based on the model's token-level uncertainty. Furthermore, \rorekv integrates these mechanisms into vLLM through fused CUDA kernels that preserve its paged KV layout and CUDA graph execution, improving reasoning accuracy while delivering end-to-end serving speedups compared with several evaluated state-of-the-art methods.

\section{Conclusion}
\label{sec:6-conc}

We presented \rorekv{}, a KV cache compression method for large reasoning models. \rorekv{} combines uncertainty-guided adaptive pruning with redundancy-aware token scoring to use the attention budget more effectively. It keeps more context when the model is uncertain, prunes more when it is confident, and gives less attention to repeated content.  It also removes redundant tokens in a lightweight scheme to give less attention to repeated content, so that the available attention budget is used for more useful information. Across three reasoning benchmarks and two models, \rorekv{} consistently outperforms all evaluated state-of-the-art compression methods under tight budgets. We also developed a drop-in \vllm{} plugin with fused CUDA decode kernels that preserves paged cache management, prefix caching, and CUDA graph execution. This integration turns reduced attention traffic into end-to-end serving improvements. When serving DeepSeek-R1-Distill-Llama-8B in \vllm{}, \rorekv{} achieves a 1.27$\times$ speedup and 38\% higher throughput on \aime{} as the most complex evaluated dataset, and a 1.43$\times$ speedup and 45\% higher throughput on the long synthetic benchmark, compared with \fullkv{}. These results show that adapting compression to reasoning behavior can improve both accuracy and serving efficiency.
\section*{Acknowledgment}
This material is based upon work supported in part by CoCoSys, one of seven centers in JUMP 2.0, a Semiconductor Research Corporation (SRC) program sponsored by DARPA, as well as the National Science Foundation (NSF) under Grant Number CNS-2420977. 
Disclaimer: Any opinions, findings, conclusions, or recommendations expressed in this material are those of the authors and do not necessarily reflect the views of the National Science Foundation.

\bibliographystyle{mlsys2025}
\bibliography{ref-main}

\clearpage
\appendix

\section{Thought-level and hybrid-level pruning}
\label{app:thought} 
As an alternative to token-level, we explored \textit{thought-level} pruning, which segments reasoning traces into coherent steps (roughly one per sentence or paragraph), constructs compact semantic representations, and dynamically restricts attention to a small subset of relevant past thoughts at each decoding step. This approach aims to capture semantic redundancy across reasoning steps, such as repeated conclusions and revisited subproblems, that token-level attention scores may overlook due to their sensitivity to local and positional information. We also evaluate hybrid-level when we combine both token-level uncertainty and thought-level redundancy together to see how it performs. Overall, practical challenges and comparative evaluation ultimately led us to retain the token-level approach. 
\subsection{Thought-Level Pruning Algorithm}
\label{app:thought:alg}
The approach layers thought-level selection on top of a token-level sparse-attention
backend. Each completed \emph{thought} corresponds to a contiguous token range
$[\text{start}, \text{end})$ in the KV cache. During generation, we detect thought
boundaries in the output stream using paragraph separators (for reasoning-style
models). When a boundary is detected, we encode the completed thought's decoded text
using a sentence embedder and store the resulting $L_2$-normalized representation
$\mathbf{e} \in \mathbb{R}^d$. We designate the first completed thought $t_0$ (the problem statement) as the \emph{anchor}, which is always retained. Let $\mathcal{B} = (b_1, \dots, b_n)$ be the ordered
list of completed non-anchor thoughts , from oldest ($b_1$) to newest ($b_n$), let
$\mathbf{q}$ be a query derived from recent reasoning context, and let
$\tau \in [0,1]$ be a redundancy threshold. The selected set of thoughts is
\begin{equation}
\resizebox{0.88\columnwidth}{!}{$\displaystyle
\mathcal{S} =
\underbrace{\{t_0\}}_{\text{anchor}}
\;\cup\;
\underbrace{\{b_{n-W+1}, \ldots, b_n\}}_{
    \text{recent window: last } W \text{ thoughts}
}
\;\cup\;
\underbrace{\operatorname{TopK}_{\mathrm{dedup}},}_{
    \text{retrieved from } \{b_1, \ldots, b_{n-W}\}
}
$}
\label{eq:thought-selection}
\end{equation}
where $t_0$ is the problem statement (always kept), $W$ is the number of most
recently completed thoughts retained, and the retrieved bucket contains up to $K$
additional thoughts selected from the older ones using the score defined below. $\operatorname{TopK}_{\mathrm{dedup}}$ denotes redundancy-aware top-$K$
selection: candidates are considered in descending order of score and a candidate
is skipped if its similarity to an already-selected thought exceeds $\tau$, so that
near-duplicate thoughts do not both consume the budget. The
retrieval score is

\begin{equation}
\resizebox{0.7\columnwidth}{!}{$\displaystyle
\mathrm{score}(b_j)
=
\cos\!\big(\mathbf{q}, \mathbf{e}_{b_j}\big)
\;\cdot\;
\mathrm{e}^{-\alpha\,(n-j)}.
$}
\label{eq:retrieval-score}
\end{equation}

The score combines semantic similarity to the query with a preference for more recent thoughts. Here, $\alpha$ controls the strength of the recency decay, and $n-j$ indicates how many thoughts are newer than candidate $b_j$. We rank the older thoughts by their scores and consider them in descending order. To avoid keeping redundant information, we skip any candidate whose similarity to an already-selected thought exceeds $\tau$. This continues until $K$ additional thoughts have been selected or no candidates remain. Finally, attention is restricted to the token ranges of the selected thoughts and the currently open thought. The token-level sparse-attention backend then applies its usual top-$k$ selection within these allowed tokens.

\subsection{Experiments}
\label{app:thought:exp}

\paragraph{Setup.}
We compare thought-level redundancy removal against the token-level redundancy
method of the main paper and later with hybrid-level, where we combine thought-level redundancy-aware and token-level uncertainty-guided adaptive pruning.

All methods are built on \rocketkv as the backend and share the same token budget
(i.e., 256); they differ only in the redundancy-removal stage. For the
thought-level variant, thoughts are segmented on paragraph boundaries, embedded with
a sentence-transformer (\texttt{all-MiniLM-L6-v2}), and selected with anchor pinning,
recent window $W$, and retrieval budget $K$ (denoted \texttt{top-t}). For the
token-level variant, we use the main-paper configuration. We report two settings: redundancy
removal alone, and a hybrid method that additionally applies uncertainty-guided adaptive pruning at the token-level and redundancy-aware in thought-level. In the latter, the uncertainty thresholds $\tau_{\text{low}}$ and $\tau_{\text{high}}$ are held identical across both methods for each dataset/model pair,
so the only variable is the redundancy-removal granularity. Accuracy is exact-match
on the final answer.
Unlike the main paper, we use $n=1$ for pass@1 for the \aime dataset.

\paragraph{Redundancy removal alone.}
Table~\ref{tab:thought:redonly} compares the two levels with redundancy
removal only. Token-level removal is better on both datasets: 
\gsm (80.59\% vs.\ 78.09\%)  and \aime (6.67\% vs.\ 3.33\%).

\begin{table}[h]
\centering
\caption{Accuracy of redundancy removal only (without uncertainty-guided adaptive pruning) using \deepqwen under a 256-token KV cache budget for \gsm  and \aime benchmarks.}
\label{tab:thought:redonly}
\begin{tabular}{llcc}
\toprule
Dataset & Level & Accuracy \\
\midrule
\aime  & Token-level              & \textbf{6.67\%} \\
\aime  & Thought-level ($W{=}4, K{=}8$) & 3.33\% \\
\midrule
\gsm   & Token-level              & \textbf{80.59\%} \\
\gsm   & Thought-level ($W{=}4, K{=}8$) & 78.09\% \\
\bottomrule
\end{tabular}
\end{table}

\paragraph{Thought-level hyperparameters sweep.}
To rule out an unlucky configuration, we swept the thought-level recent window $W$ and
retrieval budget $K$ on \gsm. The results are shown in Table~\ref{tab:thought:sweep}. Accuracy is
insensitive to $W$ and varies negligibly with $K$, peaking at
$K = 8$ with 78.70\%. Even this best thought-level setting remains below the
token-level result of 80.59\%, confirming the gap is not a tuning artifact.

\begin{table}[h]
\centering
\caption{Thought-level hyperparameter sweep on \gsm over \deepqwen, token KV cache budget 256. The accuracy of the token-level baseline is 80.59\%.}
\label{tab:thought:sweep}
\begin{tabular}{lccc}
\toprule
 & $K{=}4$ & $K{=}8$ & $K{=}16$ \\
\midrule
$W{=}2$  & 77.63\% & \textbf{78.70\%} & 78.01\% \\
$W{=}4$  & 77.63\% & \textbf{78.70\%} & 78.01\% \\
$W{=}8$  & 77.63\% & \textbf{78.70\%} & 78.01\% \\
\bottomrule
\end{tabular}
\end{table}

\paragraph{Hybrid-level.}
We next add the second component of the main method in token-level, uncertainty-guided adaptive
pruning, keeping $(\tau_{\text{low}}, \tau_{\text{high}})$ fixed across both levels for
each dataset/model pair. Table~\ref{tab:thought:full} reports the results across both
models and all three datasets. Token-level removal is best in every row except on \mathf over \deepqwen, where even in this case the gap is marginal (i.e. 0.2\%).
On \aime the gap is large and
consistent: token-level reaches 26.67\% (\deepllama) and 23.33\% (\deepqwen), whereas the
best thought-level configuration reaches only 13.33\% and 16.67\%, respectively. 

Beyond accuracy, the thought-level variant carries
practical costs that token-level avoids: it depends on an external sentence embedder
in the decoding loop, adds per-thought embedding latency, and is sensitive to how the
trace is segmented into thoughts, with accuracy depending on representation and boundary
choices that are not robust across tasks and models. Token-level removal, by contrast,
operates directly on the quantities that govern attention, requires no auxiliary model,
and remains robust without per-task tuning. 

Because token-level  is substantially
better on the hardest benchmark, and is simpler and cheaper to deploy, we adopt the
token-level method and retain the thought-level and hybrid-level approach here to show our exploration and findings.

\begin{table}[!ht]
\centering
\caption{Hybrid-level pruning, token KV cache budget 256.
Token-level uses $\lambda=0.1$, redundancy window 8; thought-level uses $K{=}8$ with the
indicated window $W$. Uncertainty thresholds $(\tau_{\text{low}}, \tau_{\text{high}})$ are
identical across levels within each block.}
\label{tab:thought:full}
\resizebox{0.5\textwidth}{!}{%
\begin{tabular}{lllc}
\toprule
Model & Dataset $(\tau_{\text{low}}, \tau_{\text{high}})$ & Method & Accuracy \\
\midrule
\multirow{5}{*}{\deepllama} & \multirow{5}{*}{\aime ($0.0$--$0.7$)}
 & Token-level                 & \textbf{26.67\%} \\
 & & Hybrid-level ($W{=}4$)    & 6.67\% \\
 & & Hybrid-level ($W{=}8$)    & 13.33\% \\
 & & Hybrid-level ($W{=}16$)   & 6.67\% \\
 & & Hybrid-level ($W{=}32$)   & 10.00\% \\
\midrule
\multirow{5}{*}{\deepqwen} & \multirow{5}{*}{\aime ($0.0$--$0.6$)}
 & Token-level                 & \textbf{23.33\%} \\
 & & Hybrid-level ($W{=}4$)    & 16.67\% \\
 & & Hybrid-level ($W{=}8$)    & 6.67\% \\
 & & Hybrid-level ($W{=}16$)   & 16.67\% \\
 & & Hybrid-level ($W{=}32$)   & 16.67\% \\
\midrule
\multirow{3}{*}{\deepllama} & \multirow{3}{*}{\gsm ($0.0$--$0.7$)}
 & Token-level                 & \textbf{75.28\%} \\
 & & Hybrid-level ($W{=}8$)    & 73.62\% \\
 & & Hybrid-level ($W{=}32$)   & 74.83\% \\
\midrule
\multirow{3}{*}{\deepqwen} & \multirow{3}{*}{\gsm ($0.1$--$0.6$)}
 & Token-level                 & \textbf{81.43\%} \\
 & & Hybrid-level ($W{=}8$)    & 80.06\% \\
 & & Hybrid-level ($W{=}32$)   & 80.14\% \\
\midrule
\multirow{3}{*}{\deepqwen} & \multirow{3}{*}{\mathf ($0.0$--$0.4$)}
 & Token-level                 & 86.80\% \\
 & & Hybrid-level ($W{=}8$)    & 80.20\% \\
 & & Hybrid-level ($W{=}32$)   & \textbf{87.00\%} \\
\bottomrule
\end{tabular}
}
\end{table}

\section{Adaptive layer-wise token budget allocation}
We also implemented layer-wise token budget allocation optimization. Not all attention heads and layers contribute equally during decoding. Some layers consistently capture more useful information, while others have a limited impact. Allocating the same KV budget across all layers is therefore inefficient. To address this, inspired by HeadkV-R~\cite{headkvr22025}, we introduce an adaptive token budget allocation mechanism that redistributes the KV cache budget based on head importance and integrates it with \rocketkv.

To estimate head importance, we use the concept of \textit{``retrieval heads''} introduced by Wu et al.~\cite{retrievalHead2024}. Retrieval heads are attention heads that play an important role in retrieving relevant information from long contexts. The \textit{``retrieval head''} work identifies these heads using the \needle test~\cite{needle2023}. For each model, the test places a specific ``needle'' fact inside a long context and measures which heads assign higher attention scores to the needle tokens. Heads that consistently attend to the needle are considered more important for retrieval. Since retrieval heads are mainly model-specific, we compute the retrieval-head scores once for each model and reuse them across all datasets. This gives us an importance table for all heads, which we later use to guide the KV cache budget allocation.

Our method follows importance-based budget allocation and layer-wise top-$k$ selection during decoding.
To do so, we estimate the importance of each attention head using precomputed scores. Let $w_{l,h}$ denote the normalized importance of head $h$ in layer $l$, where
\begin{equation}
\sum_{l,h} w_{l,h} = 1.
\end{equation}
Given a total token budget $B$, we distribute capacity across heads by allocating a non-uniform budget:
\begin{equation}
b_{l,h} \propto w_{l,h},
\end{equation}
subject to a minimum allocation per head to ensure stability. This produces a per-head capacity matrix $b \in \mathbb{R}^{L \times H}$, where $L$ and $H$ denote the number of layers and heads, respectively. While per-head budgets are desirable, they are not fully compatible with high-performance attention implementations such as \vllm and FlashAttention, which assume a uniform token layout. To maintain efficiency, we aggregate the per-head capacities into a single layer-wise budget:
\begin{equation}
k_l = \frac{1}{H} \sum_{h=1}^{H} b_{l,h},
\end{equation}
where $k_l$ is the effective top-$k$ budget for layer $l$. 

\textbf{Decode-time selection.}
During decoding, we replace the fixed top-$k$ in \rocketkv with the adaptive budget $k_l$. For each layer $l$, token selection is performed as:
\begin{equation}
\mathcal{I}_l = \text{top-$k$}\!\left(\{\tilde{s}_{t,i}\}_i,\ k_l\right),
\end{equation}
where $\tilde{s}_{t,i}$ denotes the adjusted attention score (including redundancy-aware and uncertainty-guided modifications).
The selected tokens $\mathcal{I}_l$ are then used for exact attention computation.

Table~\ref{tab:layerwise} shows the corresponding results of layer-wise implementation in \rorekv when the token budget is 512 compared to \fullkv on \gsm. 
For all of these results, the observation window is 32, the temperature is 1, and \topp is 1.0, which is different from the setting we have in the main paper. As it is shown, for \deepqwen, this approach even outperforms \fullkv.

\begin{table}[!ht]
\centering
\caption{Accuracy of different models under a 512-token KV cache budget for \gsm benchmarks.} 
\label{tab:layerwise}
\begin{tabular}{llcc}
\toprule
\textbf{Dataset} & \textbf{Model} & \textbf{Token Budget} & \textbf{Accuracy} \\
\midrule
\gsm   & \deepllama & \fullkv & 79.23\% \\
\gsm   & \deepllama & 512 & 79.15\% \\
\gsm   & \deepqwen & \fullkv & 85.22\% \\
\gsm   & \deepqwen & 512 & 85.75\% \\

\bottomrule
\end{tabular}
\end{table}

Integration with \vllm causes some throughput degradation, due to challenges related to KV cache layout and scheduling. 
Therefore, we leave this optimization for future works.

\section{Implementation Details of \vllm Integration}
\label{app:integration}

This appendix expands on the kernel-level mechanics summarized in
Section~\ref{sec:vllm-integration}. For each kernel launch, the
mapping from logical token positions to physical pages (the block table)
is loaded once into shared memory at kernel entry, so subsequent loads
issue against contiguous addresses, eliminating the per-token block-table
lookup that a naive port of a contiguous-KV kernel would require.

\paragraph{Terminology.}
We refer to several CUDA-specific concepts. A \emph{thread block} (also called a Cooperative Thread
Array, or CTA) is a
group of threads that runs together on a single streaming multiprocessor
and shares a software-managed scratchpad. A \emph{warp} is a group of 32
threads that execute in lockstep on NVIDIA GPUs, and a \emph{lane} is one
thread within a warp. CUDA graphs are efficient ways to execute a series of kernels by recording a series of kernel launches and replaying them on the GPU. CUDA graphs are particularly useful for cases where kernels exit faster than the CPU can issue them; the only drawback is that there can be no operations that result in a CPU-GPU sync within the graph. CUDA graphs are heavily used in vLLM, particularly its asynchronous scheduler, to keep the GPU constantly busy while the CPU waits on results.

\subsection{Phase A: page-level approximate scoring}
\label{app:phase-a}

A FP32 non-WMMA matmul of $Q$ against
$\bar K$ produces per-(page, GQA-rep) approximate logits (we observed that using WMMA resulted in overhead that wasn't the most efficient for our use case), fused with a running sum/max accumulator for
each query head (the actual softmax reduction is deferred to the Phase
A reduction kernel) that uses \texttt{exp2f} on the special-function
unit and warp-shuffle reductions across threads. The kernel is templated on
$(\mathrm{TILE\_SIZE}, R, \mathrm{dtype})$ with 60 explicit instantiations
so that every inner loop has compile-time bounds and unrolls fully;
runtime dispatch is a flat nested \texttt{switch}, never a branch inside
the kernel body. 

\subsection{Fused redundancy scoring}
\label{app:redundancy}

The redundancy score $r_{t,i}$ from Equation~\ref{redundancy-eq} is
computed inside the same Phase A kernel as the approximate logits. Per
tile, we
\begin{enumerate}
\item stream raw keys for $\mathrm{TILE\_SIZE}$
      positions into a shared-memory buffer;
\item compute $1/\|\mathbf{k}_i\|_2$ per position via per-warp shuffle
      reductions;
\item run a non-WMMA matmul with the unit-normalized centroid
      $\hat{\boldsymbol{\mu}}^{(w)}_t$ from Equation~\ref{centroid-eq}
      (broadcast across a padded $16 \times \mathrm{HEAD\_DIM}$ row)
      against the raw keys to obtain
      $\langle \hat{\boldsymbol{\mu}}^{(w)}_t, \mathbf{k}_i \rangle$;
\item write the per-position cosine similarities into a global
      \texttt{redundancy\_scores[B, H\_kv, len\_k]} buffer.
\end{enumerate}

The Phase A reduce kernel later
applies the redundancy multiplier $(1 - \lambda r_{t,i})$ from
Equation~\ref{s-tild} per position when scattering scores. The net cost
is one additional WMMA and one extra global write per tile, with no
additional pass over the KV cache.

\subsection{Phase B: top-$k$ selection}
\label{app:topk}

Standard top-$k$ kernels degrade under dynamic shapes during CUDA graph
capture, falling back to generic implementations with poor memory
behavior. We instead use an on-chip bucket scheme inspired by approximate
top-$k$~\cite{faiss2019}. Each thread maintains a per-thread bucket of
size $P+1$ (with $P=4$) using a branchless descending bubble that compiles
to predicated selects, so there is no warp divergence and no early exit.
After the streaming pass each thread holds its top-$P$ candidates, after
which we issue a single \texttt{cub::BlockRadixSort::SortDescending} so
that the on-chip top-$k$ list is globally sorted before being written
out. Fusing the sort here saves a separate sort kernel launch and a
global round-trip, and it is what allows Phase C to consume the prefix
$[0, k_t)$ directly under per-request adaptive budgets. Templates, a particular instantiation of a kernel with certain parameters, round the requested $k$ so that the block size
$k_{\text{kernel}}/P$ stays a compile-time constant. \vllm scheduling
pads any unused slots in the top-$k$ output to index $0$, which Phase C
tolerates.

\subsection{Phase C: tiled exact sparse attention}
\label{app:phase-c}

Phase C reads the sorted index list and processes 16 selected positions
at a time (one \emph{tile}). Keys and values are loaded with 128-bit
vectorized accesses, which on bf16 or fp16 transfers 8 elements per
instruction. Four threads cooperate per position in the $QK$ dot product,
reducing across the head dimension via warp shuffles, and the per-tile
online-softmax max and sum are reduced cooperatively across 16 lanes
within the warp. This eliminates the usual thread-0 bottleneck before
the result is fused into the running $V$ accumulator.

We use a \emph{boundary peeling} pattern to handle the final tile of the
loop. Because the per-request budget $k_t$ may not be a multiple of the
tile size, the last tile of a request may contain fewer valid positions
than the tile is sized for, and may extend past $k_t$ if the kernel was
launched with an upper bound $k_{\max} > k_t$. We peel this last tile out
of the main loop and process it with a separate code path that takes the
runtime tile size as input, and that masks any position beyond $k_t$ by
assigning it a score of $-\infty$ before the softmax, which makes its
softmax weight zero and removes its contribution from the attention
output. All preceding tiles run with the tile size as a compile-time
constant, so their inner loops fully unroll. Processing the boundary
tile separately rather than inside the main loop avoids per-iteration
runtime checks in the common case. Moreover, to accelerate loading of tokens, we use \texttt{cp.async} with double buffering. This allows us to overlap loading of the next tile of tokens with compute of the current tile.

\subsection{Per-request adaptive budgets}
\label{app:per-request}

Each request carries its own adaptive budget $k_t$ from
Section~\ref{sec:vllm-integration}. 
We pass these to the kernels as a 32-bit
integer device tensor \texttt{topk\_per\_req[B]}. Phases A and B size
their work and shared-memory buffers using a launch-time upper bound
$k_{\max}$. Phase C reads each request's budget at runtime, shrinks its
tile loop accordingly, and applies the boundary mask described above.
Because the Phase B output is descending sorted, the prefix of length
$k_t$ is exactly that request's approximate top-$k_t$, so no re-sort and no
per-request kernel launch is needed. The CPU never reads any tensor
value during decoding, which is what makes the entire pipeline
capturable inside a single CUDA graph.

\subsection{CUDA graph compatibility and short-circuits}
\label{app:cuda-graphs}

Several design choices follow from the requirement to be capturable in a
single CUDA graph. Attention metadata, such as sequence lengths, masks,
and block tables, is precomputed once per decoding step and reused across
layers. Kernel execution paths avoid request-dependent CPU branching: prefill requests early-exit per thread block, and padded
requests with \texttt{seq\_len=0} take a fast path that writes neutral
partials (max of $-\infty$, sum of $0$) so that the reduction kernel
produces zeros without touching the cache. Furthermore, the kernels
themselves early-exit per thread block when assigned to a
prefill request or when $k_t \geq \mathrm{seq\_len}$ for that request,
which handles within-batch heterogeneity in sequence length without a
separate launch.

\subsection{Split-KV work distribution}
\label{app:split-map}

Section~\ref{sec:vllm-integration} introduced split-KV execution, in which
each request's cached sequence is cut into $S$ segments that are processed
by separate thread blocks and merged by a reduction kernel. This section
describes in details how Phase A decides how many segments each request receives.

Phase A streams every cached token of a request, so its work per request
is proportional to that request's cached length, and the lengths within
one decode batch can differ by a hundredfold or more. The usual strategy,
used by FlashAttention-2, picks one split
count for the whole batch from the batch shape alone, so every request
receives the same number of thread blocks. This is wasteful whenever the
batch is skewed. A common case in long reasoning workloads: several long
requests are running, one of them finishes, and the \vllm scheduler fills
the freed KV cache with several new short prompts. The thread blocks
assigned to the short requests exit almost immediately, the long requests
still have to be streamed by the same small number of blocks each, and
the GPU idles until they finish and Phase B can start.

We therefore assign thread blocks in proportion to work. Once per decode
step, the attention metadata builder, which \vllm runs on the CPU outside
CUDA graph capture, plans how many splits each request receives and writes
the plan into two small device tensors: a table with one entry per thread
block giving (request, split index, number of splits of that request), and
a table giving each request's first entry. Phase A is launched with a
fixed grid, one block per table entry for each KV head, and each block
reads its entry to learn which request and which segment it owns. Because
the grid size depends only on the batch size and never on the sequence
lengths, a CUDA graph captured for a batch size can be replayed every step
while the table contents change underneath it. The plan is computed once
per step and shared by all layers, since cached lengths do not change
between the layers of one step.

\paragraph{Greedy planning.}
Every request first receives one split. The remaining entries are handed
out one at a time, always to the request whose blocks currently have the
most tiles to process. The handout runs in
three passes with a progressively looser cap on splits per request. The
first pass only creates splits with at least four tiles of work, because
splitting a tiny request only duplicates per-block setup (loading the
query, the block table and the recent-context centroid). The second pass
allows splits down to one tile. The third adds empty padding splits so
that every launched entry has an owner; a block that receives one writes
a neutral partial result (maximum $-\infty$, sum $0$) and exits. The plan
takes a few microseconds of CPU time per step.

Phase C splits along the sequence axis as well, but its work per request
is bounded by the budget $k_t$ rather than by the cached length, so it
uses a single split count for the batch, derived from the batch shape and
the budget, and does not read the Phase A table. Phase B does not split.

\section{Limitations and Future Work}
\label{sec:5-limit}

While \rorekv\ achieves strong accuracy and high throughput under tight KV budgets, there are several areas for further improvement. 
Although we provide an efficient CUDA implementation and integrate it into \vllm, there is still room to further optimize throughput with additional system-level engineering, such as more aggressive kernel tuning and scheduling strategies.  Hopper-class hardware
features such as the Tensor Memory Accelerator can replace
our manual vectorized loads for structured access patterns. 
Extending support to other inference frameworks such as SGLang is left for future work. 
Finally, \rorekv, like many approaches, relies on a small set of hyperparameters, including uncertainty thresholds and redundancy coefficients. While these are stable across our experiments, reducing or eliminating such dependencies would further improve robustness and ease of deployment.

\end{document}